\documentclass[10pt,conference]{IEEEtran}
\ifCLASSINFOpdf
\else
\fi
\usepackage{booktabs}
\usepackage{amssymb,amsmath,amsthm, amsfonts}
\usepackage[ruled,vlined,linesnumbered]{algorithm2e}

\usepackage{multirow}
\usepackage{makecell}
\usepackage{subcaption}
\usepackage{graphicx}
\usepackage{adjustbox}
\usepackage{framed}
\usepackage{hyperref}
\begin{document}
\title{Adversarial Sample Detection for Deep Neural Network through Model Mutation Testing}

\author{
% anonymous
\IEEEauthorblockN{Jingyi Wang\IEEEauthorrefmark{2},
Guoliang Dong\IEEEauthorrefmark{3},
Jun Sun\IEEEauthorrefmark{2},
Xinyu Wang\IEEEauthorrefmark{3},
Peixin Zhang\IEEEauthorrefmark{3}
}
% \IEEEauthorblockA{\IEEEauthorrefmark{1}College of Computer Science and Software Engineering, Shenzhen University}
\IEEEauthorblockA{\IEEEauthorrefmark{2}Singapore University of Technology and Design}
\IEEEauthorblockA{\IEEEauthorrefmark{3}Zhejiang University}

}

% \IEEEspecialpapernotice{(Invited Paper)}

\maketitle

\begin{abstract}
Deep neural networks (DNN) have been shown to be useful in a wide range of applications. However, they are also known to be vulnerable to adversarial samples. By transforming a normal sample with some carefully crafted human imperceptible perturbations, even highly accurate DNN make wrong decisions. Multiple defense mechanisms have been proposed which aim to hinder the generation of such adversarial samples. However, a recent work show that most of them are ineffective. In this work, we propose an alternative approach to detect adversarial samples at runtime. Our main observation is that adversarial samples are much more sensitive than normal samples if we impose random mutations on the DNN. We thus first propose a measure of `sensitivity' and show empirically that normal samples and adversarial samples have distinguishable sensitivity. We then integrate statistical hypothesis testing and model mutation testing to check whether an input sample is likely to be normal or adversarial at runtime by measuring its sensitivity. We evaluated our approach on the MNIST and CIFAR10 datasets. The results show that our approach detects adversarial samples generated by state-of-the-art attacking methods efficiently and accurately.              
\end{abstract}

\section{Introduction}\label{sec:intr}

In recent years, deep neural networks (DNN) have been shown to be useful in a wide range of applications including computer vision~\cite{he2016deep}, speech recognition~\cite{xiong2016achieving}, and malware detection~\cite{yuan2014droid}. However, recent research has shown that DNN can be easily fooled~\cite{Szegedy2013Intriguing,FGSM} by adversarial samples, i.e., normal samples imposed with small, human imperceptible changes (a.k.a.~perturbations). Many DNN-based systems like image classification~\cite{DeepFool,JSMA,CW,xiao2018spatially} and speech recognition~\cite{carlini2018audio} are shown to be vulnerable to such adversarial samples. This undermines using DNN in safety critical applications like self-driving cars~\cite{bojarski2016end} and malware detection~\cite{yuan2014droid}.

To mitigate the threat of adversarial samples, the machine learning community has proposed multiple approaches to improve the robustness of the DNN model. For example, an intuitive approach is data augmentation. The basic idea is to include adversarial samples into the training data and re-train the DNN~\cite{shaham2015understanding,kurakin2016adversarial,deeptest}. It has been shown that data augmentation improves the DNN to some extent. However, it does not help defend against unseen adversarial samples, especially those obtained through different attacking methods. Alternative approaches include robust optimization and adversarial training~\cite{sinha2018certifying,tramer2017ensemble,yu2018towards,madry2017towards}, which take adversarial perturbation into consideration and solve the robust optimization problem directly during model training. However, such approaches usually increase the training cost significantly.    

Meanwhile, the software engineering community attempts to tackle the problem using techniques like software testing and verification. In~\cite{deeptest}, neuron coverage was first proposed to be a criteria for testing DNN. Subsequently, multiple testing metrics based on the range coverage of neurons were proposed~\cite{ma2018deepgauge}. Both white-box testing~\cite{deepxplore}, black-box testing~\cite{deeptest} and concolic testing~\cite{sun2018concolic} strategies have been proposed to generate adversarial samples for adversarial training. However, testing alone does not help in improving the robustness of DNN, nor does it provide guarantee that a well-tested DNN is robust against new adversarial samples. The alternative approach is to formally verify that a given DNN is robust (or satisfies certain related properties) using techniques like SMT solving~\cite{reluplex,weng2018towards} and abstract interpretation~\cite{ai2}. However, these techniques usually have non-negligible cost and only work for a limited class of DNN (and properties).

\begin{figure*}[t]
\centering
\begin{subfigure}[b]{0.45\textwidth}
   \centering 
   \includegraphics[height=1in]{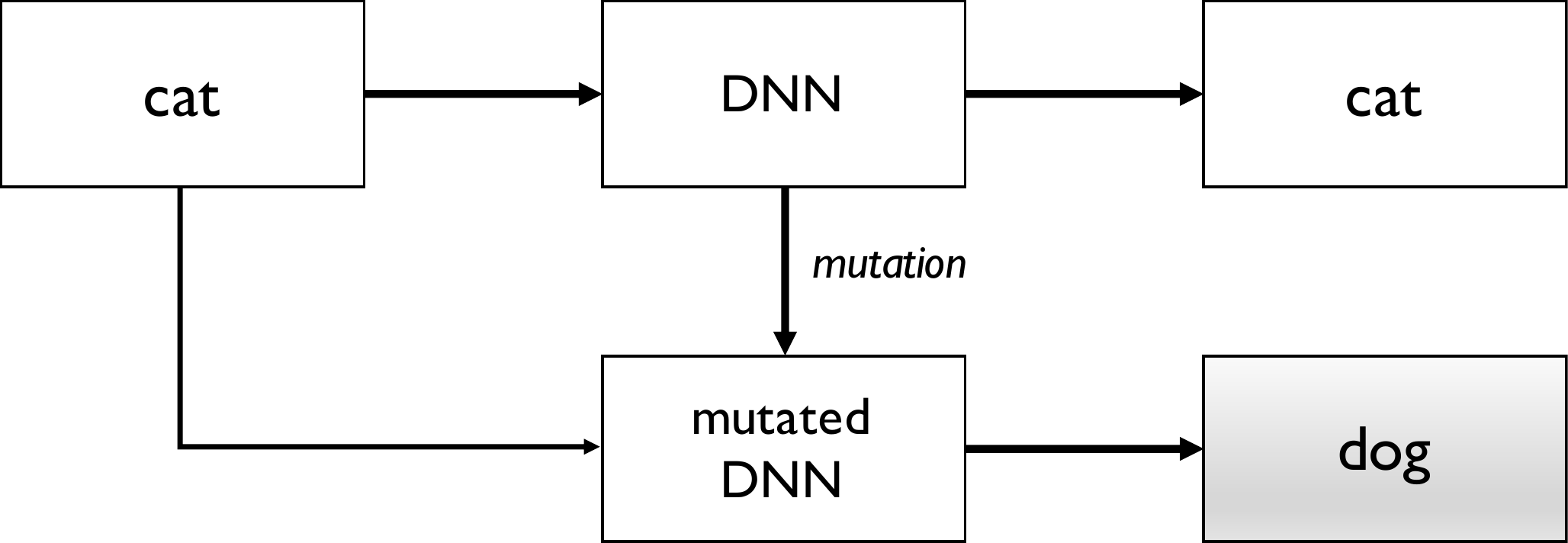}
   % \caption{NASA Logo} 
   \label{fig:lcr:nor}
\end{subfigure}% 
\begin{subfigure}[b]{0.45\textwidth}
   \centering 
   \includegraphics[height=1in]{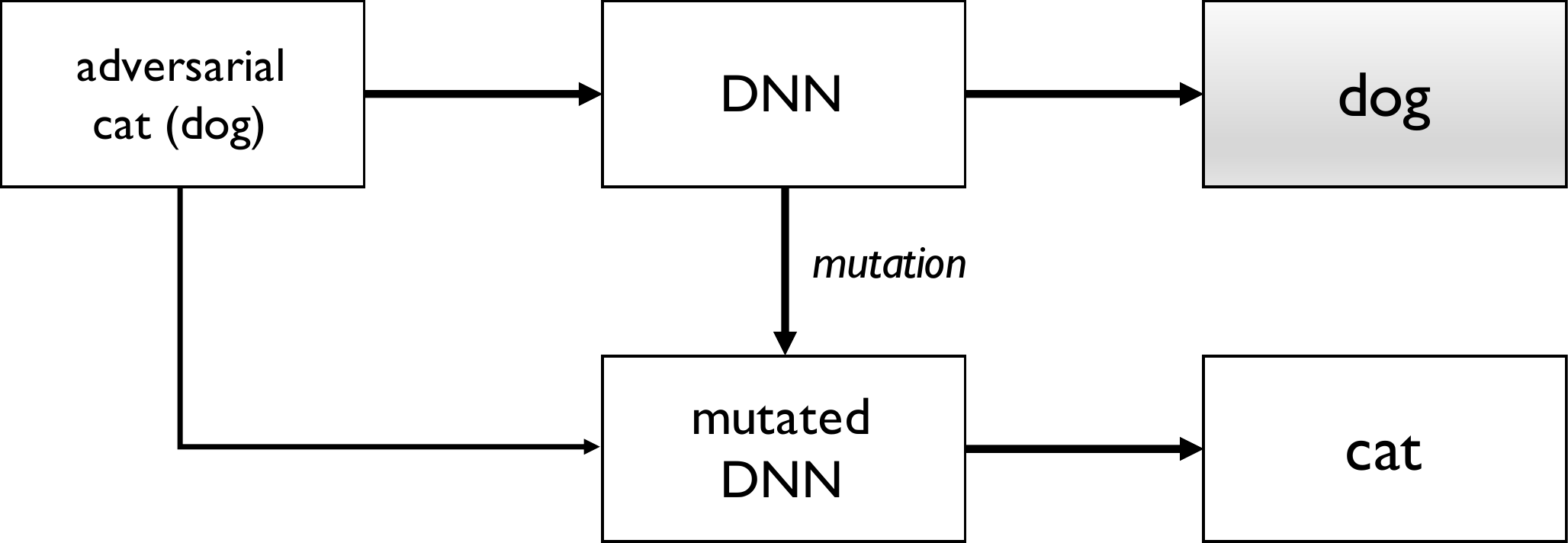}
   % \caption{Orion Logo} 
   \label{fig:lcr:adv}
\end{subfigure}%
   \caption{Label change of a normal sample and an adversarial sample against DNN mutation models.}
   \label{fig:lcr}
\end{figure*}

In this work, we provide a complementary perspective and propose an approach for detecting adversarial samples at runtime. The idea is that, given an arbitrary input sample to a DNN, to decide at runtime whether it is likely to be an adversarial sample or not. If it is, we raise an alarm and report that the sample is `suspicious' with certain confidence. Once detected, it can be rejected or checked depending on different applications. Our detection algorithm integrates mutation testing of DNN models~\cite{deepmutation} and statistical hypothesis testing~\cite{sprt}. It is designed based on the observation that adversarial samples are much more sensitive to mutation on the DNN than normal samples, i.e., if we mutate the DNN slightly, the mutated DNN is more likely to change the label on the adversarial sample than that on the normal one. This is illustrated in Figure~\ref{fig:lcr}. The left figure shows a label change on a normal sample, i.e., given a normal sample which is  classified as a cat, a label change occurs if the mutated DNN classifies the input as a dog. The right figure shows a label change on an adversarial sample, i.e., given an adversarial sample which is mis-classified as a dog, a label change occurs if the mutated DNN classifies the input as a cat. Our empirical study confirms that the label change rate (LCR) of adversarial samples is significantly higher than that of normal samples against a set of DNN mutants. We thus propose a measure of a sample's sensitivity against a set of DNN mutants in terms of LCR. We further adopt statistical analysis methods like receiver operating characteristic (ROC~\cite{roc}) to show that we can distinguish adversarial samples and normal samples with high accuracy based on LCR. Our algorithm then takes a DNN model as input, generates a set of DNN mutants, and applies statistical hypothesis testing to check whether the given input sample has a high LCR and thus is likely to be adversarial.

We implement our approach as a self-contained toolkit called mMutant~\cite{tool}. We apply our approach on the MNIST and CIFAR10 dataset against the state-of-the-art attacking methods for generating adversarial samples. The results show that our approach detects adversarial samples efficiently with high accuracy. All four DNN mutation operators we experimented with show promising results on detecting 6 groups of adversarial samples, e.g., capable of detecting most of the adversarial samples within around 150 DNN mutants. In particular, using DNN mutants generated by Neuron Activation Inverse (NAI) operator, we manage to detect 96.4\% of the adversarial samples with 74.1 mutations for MNIST and 90.6\% of the adversarial samples with 86.1 mutations for CIFAR10 on average.

%On average, the four applied operators could achieve accuracy of 92.6\%/93.3\%/79.4\%/85.5\% with 105.5/111.9/177.7/149.6 mutated models for MNIST and 85.5\%/90.6\%/56.6\%/74.8\% ($\rho$=1) with 121.7/86.1/303/176.2 mutated models for CIFAR10 on detecting the 6 kinds of adversarial samples. 
% Meanwhile, we maintain high detection accuracy of normal samples as well, i.e., 92.5\%/92.7\%/95.2\%/94.5\% for MNIST ($\rho$=1.5) and xx/xx/xx/xx ($\rho$=2?) for CIFAR10 for the above 4 operators respectively.

% The rest of the paper is organized as follows. Section~\ref{sec:back} provides the necessary background of this work. Section~\ref{sec:app} presents the details of our approach. Section~\ref{sec:ev} shows our experiment results. Section~\ref{sec:re} reviews related works and Section~\ref{sec:con} concludes.

% \input{background.tex}

\section{Background}\label{sec:back}
In this section, we review state-of-the-art methods for generating adversarial samples for DNN, and define our problem.

\subsection{Adversarial Samples for Deep Neural Networks} 
%Deep learning systems are emerging in many different applications including computer vision, natural language processing, and malware detection. Different DNN architectures like Convolutional Neural networks (CNN) and Recurrent Neural Networks (RNN) have been proposed for different application scenarios. 
In this work, we focus on DNN classifiers which take a given sample and label the sample accordingly (e.g., as a certain object). In the following, we use $x$ to denote an input sample for a DNN $f$. We use $c_x$ to denote the ground-truth label of $x$. Given an input sample $x$ and a DNN $f$, we can obtain the label of the input $x$ under $f$ by performing forward propagation. $x$ is regarded as an \textit{adversarial sample} with respect to the DNN $f$ if $f(x)\neq c_x$. $x$ is regarded as a \textit{normal sample} with respect to the DNN $f$ if $f(x)= c_x$. Notice that under our definition, those samples in the training/testing dataset 
% which are 
wrongly labeled by $f$ are also adversarial samples. %To distinguish the two cases, we refer to as \textit{``wrongly labeled (WL)''} later.

%subsection{Attacking through Adversarial Samples}
Since Szegedy \emph{et al.} discoveried that neural networks are vulnerable to adversarial samples~\cite{Szegedy2013Intriguing}, many attacking methods have been developed on how to generate adversarial samples efficiently (e.g., with minimal perturbation). That is, given a normal sample $x$, an attacker aims to find a minimum perturbation $\Delta x$ which satisfies $f(x+\mathbf{\Delta}x)\neq c_x$. In the following, we briefly introduce several state-of-the-art attacking algorithms. \\

\noindent \textbf{FGSM}: The Fast Gradient Sign Method (FGSM)~\cite{FGSM} is designed based on the intuition that we can change the label of an input sample by changing its softmax value to the largest extent, which is represented by its gradient.
% explanation of the neural networks’ vulnerability that it is its linear nature that makes a neural network easy to be misled instead of the nonlinearity and overﬁtting. If the softmax score of a class linearly has to do with the input, we could just linearly modify
% the input to make the softmax score of the true label decrease.  
% With, 
The implementation of FGSM is straightforward and efficient. By simply adding up the sign of gradient of the cost function with respect to the input, we could quickly obtain a potential adversarial counterpart of a normal sample by the follow formulation:
\[
    \Hat{x}=x+{\epsilon}\mathbf{sign}(\nabla{\mathbf{J}(\mathbf{\theta},x,c_x)})
\]
, where $\mathbf{J}$ is the cost used to train the model, $\epsilon$ is the attacking step size and $\mathbf{\theta}$ are the parameters. Notice that FGSM does not guarantee that the adversarial perturbation is minimal. \\
% though it is an efficient manner. 

\noindent \textbf{JSMA}: Jacobian-based Saliency Map Attack (JSMA)~\cite{JSMA} is devised to attack a model with minimal perturbation which enables the adversarial sample to mislead the target model into classifying it with certain (attacker-desired) label. It is a greedy algorithm that changes one pixel during each iteration to increase the probability of having the target label. The idea is to calculate a saliency map based on the Jacobian matrix to model the impact that each pixel imposes on the target classification. With the saliency map, the algorithm picks the pixel which may have the most significant influence on the desired change and then increases it to the maximum value. The process is repeated until it reaches one of the stopping criteria, i.e., the number of pixels modified has reached the bound, or the target label has been achieved. Define

\begin{equation}
\left\{
    \begin{aligned}
       a_i = & \frac{\partial{\mathbf{F_t}(x)}}{\partial{X_i}}\\
       b_i = & \sum_{k \neq t}\frac{\partial{\mathbf{F_k}(x)}}{\partial{X_i}}\nonumber 
    \end{aligned}
\right.
\end{equation}

Then, the saliency map at each iteration is defined as follow:

\begin{equation}
   S(x,t)_i =  \left\{
   \begin{aligned} 
    & a_i\times\left|b_i\right| & if \; a_i>0\; and\; b_i<0 \\
    & 0 & otherwise  \\
   \end{aligned}
   \right.  \nonumber
\end{equation}

However, it is too strict to select one pixel at a time because few pixels could meet that definition. Thus, instead of picking one pixel at a time, the authors proposed to pick two
pixels to modify according to the follow objective: 
\[
\arg\max_{(p_1,p_2)} \left( \frac{\partial{\mathbf{F}_t(x)}}{\partial{x_{p_1}}} + \frac{\partial{\mathbf{F}_t(x)}}{\partial{x_{p_2}}} \right)\times \left|\sum_{i=p_1,p_2}\sum_{k \neq t} \frac{\partial{\mathbf{F}_k(x)}}{\partial{x_{i}}}    \right| 
\]
where$(p_1,p_2)$ is the candidate pair, and $t$ is the target class.

JSMA is relatively time-consuming and memory-consuming since it needs to compute the Jacobian matrix and pick out a pair from nearly $\binom{n}{2}$ candidate pairs at each iteration. \\

% Remind, there is a decreased-counterpart of the saliency map defined above, but they are  
\noindent \textbf{DeepFool}:
The idea of DeepFool (DF) is to make the normal samples cross the decision boundary with minimal perturbations~\cite{DeepFool}. The authors first deduced an iterative algorithm for binary classifiers with Tayler's Formula, and then analytically derived the solution for multi-class classifiers. The exact derivation process is complicated and thus we refer the readers to~\cite{DeepFool} for details.\\

\noindent \textbf{C\&W}:
Carlini \emph{et al.}~\cite{CW} proposed a group of attacks based on three distance metrics. The key idea is to solve an optimization problem which minimizes the perturbation imposed on the normal sample (with certain distance metric) and maximizes the probability of the target class label. 
% Note, JSMA minimizes the perturbations by restricting the number of pixels which need to be changed while C\&W does this by minimizing the total changes of all pixels. 
% They designs a series of loss functions to make the desired class's probability as large as possible. 
The objective function is as follow:
\[
    \arg\min \Delta x + c\cdot f(\hat{x},t)
\]
where $\Delta x$ is defined according to some distance metric, e.g, $L_0$, $L_2$, $L_{\infty}$, $\hat{x}=x+\Delta x$ is the clipped adversarial sample and $t$ is its target label. The idea is to devise a clip function for the adversarial sample such that the value of each pixel dose not exceed the legal range. The clip function and the best loss function according to~\cite{CW}  are shown as follows.
\begin{equation}
\begin{aligned}
clip:& \hat{x} &=& 0.5(tanh(\tilde{x})+1) \\ \nonumber
loss:& f(\hat{x},t) &=& \max(\max\{G(\hat{x})_c:c \neq t\}-G(\hat{x})_t, 0) 
\end{aligned}
\end{equation}
where $G(x)$ denotes the output vector of a model and $t$ is the target class. Readers can refer to~\cite{CW} for details. \\
% $\gamma$ is an hyperparameter which encourages the solver to ﬁnd an adversarial samples with high conﬁdence. 

\noindent \textbf{Black-Box}: All the above mentioned attacks are white-box attacks which means that the attackers require the full knowledge of the DNN model. Black-Box (BB) attack only needs to know the output of the DNN model given a certain input sample. The idea is to train a substitute model to mimic the behaviors of the target model with data augmentation. Then, it applies one of the existing attack algorithm, e.g., FGSM and JSMA, to generate adversarial samples for the substitute model. The key assumption to its success is that the adversarial samples transfer between different model architectures~\cite{Szegedy2013Intriguing,FGSM}. 

\subsection{Problem Definition}
Observing that adversarial samples are relatively easy to craft, a variety of defense mechanisms against adversarial samples have been proposed~\cite{guo2017countering,madry2017towards,xie2017mitigating,ma2018characterizing,song2017pixeldefend}, as we have briefly introduced in Section~\ref{sec:intr}. However, Athalye \emph{et al.}~\cite{athalye2018obfuscated} systematically evaluated the state-of-the-art defense mechanisms recently and showed that most of them are ineffective. Alternative defense mechanisms are thus desirable.

In this work, we take a complementary perspective and propose to detect adversarial samples at runtime using techniques from the software engineering community. \textit{The problem is: given an input sample $x$ to a deployed DNN $f$, how can we efficiently and accurately decide whether $f(x)=c_x$ (i.e., a normal sample) or not (i.e., an adversarial sample)?} If we know that $x$ is likely an adversarial sample, we could reject it or further check it to avoid bad decisions. Furthermore, can we quantify some confidence on the drawn conclusion?

\section{Our approach}\label{sec:app}
Our approach is based on the hypothesis that, in most cases adversarial samples are more `sensitive' to mutations on the DNN model than normal samples. That is, if we generate a set of slightly mutated DNN models based on the given DNN model, the mutated DNN models are more likely to label an adversarial sample with a label different from the label generated by the original DNN model, as illustrated in Figure~\ref{fig:lcr}.
%~\cite{novak2018sensitivity,detection_ours}
In other words, our approach is designed based on a measure of sensitivity for differentiating adversarial samples and normal samples. In the literature, multiple measures have been proposed to capture their differences, e.g., density estimate, model uncertainty estimate~\cite{feinman2017detecting}, and sensitivity to input perturbation~\cite{detection_ours}. Our measure however allows us to detect adversarial samples at runtime efficiently through model mutation testing.   

\subsection{Mutating Deep Neural Networks}
In order to test our hypothesis (and develop a practical algorithm), we need a systematic way of generating mutants of a given DNN model. We adopt the method developed in~\cite{deepmutation}, which is a proposal of applying mutation testing to DNN. Mutation testing~\cite{mutation_testing} is a well-known technique to evaluate the quality of a test suiteand, and thus is different from our work. The idea is to generate multiple mutations of the program under test, by applying a set of mutation operators, in order to see how many of the mutants can be killed by the test suite. The definition of the mutation operators is a core component of the technique. Given the difference between traditional software systems and DNN, mutation operators designed for traditional programs cannot be directly applied to DNN. In~\cite{deepmutation}, Ma \emph{et al.} introduced a set of mutation operators for DNN-based systems at different levels like source level (e.g., the training data and training programs) and model level (e.g., the DNN model).

In this work, we require a large group of slightly mutated models for runtime adversarial sample detection. Of all the mutation operators proposed in~\cite{deepmutation}, mutation operators defined at the source level are not considered. The reason is that we would need to train the mutated models from scratch which is often time-consuming. We thus focus on the model-level operators, which modify the original model directly to obtain mutated models without training. 
% The operators at model level are clearly defined in the original paper~\cite{deepmutation},and one can quickly get it if he or she refers it. 
Specifically, we adopt four of the eight defined operators from~\cite{deepmutation} shown in Table~\ref{tb:mu}. For example, NAI means that we change the activation status of a certain number of neurons in the original model. Notice that the other four operators defined in~\cite{deepmutation} are not applicable due to the specific architecture of the deep learning models we focus on in this work. 

\subsection{Evaluating Our Hypothesis}

\begin{table}[t]
\caption{DNN model mutation operators}
\centering
\label{tb:mu}
\begin{adjustbox}{width=.5\textwidth}
\begin{tabular}{@{}lll@{}}
\toprule
Mutation Operator               & Level  & Description                               \\ \midrule
Gaussian Fuzzing (GF)           & Weight & Fuzz weight by Gaussian Distribution      \\
Weight Shuffling (WS)           & Neuron & Shuffle selected weights                  \\
Neuron Switch (NS)              & Neuron & Switch two neurons within a layer  \\
Neuron Activation Inverse (NAI) & Neuron & Change the activation status of a neuron  \\ \bottomrule
\end{tabular}
\end{adjustbox}
\end{table}

%We first present the different effects of mutation testing on  
We first conduct experiments to measure the label change rate (LCR) of adversarial samples and normal samples when we feed them into a set of mutated DNN models.
Given an input sample $x$ (either normal or adversarial) and a DNN model $f$, we first adopt the model mutation operators shown in Table~\ref{tb:mu} to obtain a set of mutated models. Note that some of the resultant mutated models may be of low quality, i.e., their classification accuracy on the test data drops significantly. We discharge those low quality ones and only keep those \emph{accurate mutated models} which retain an accuracy on the test data, i.e., at least 90\% of the accuracy of the original model, to ensure that the decision boundary does not perturb too much. Once we obtain such a set of mutated models $F$, we then obtain the label $f_i(x)$ of the input sample $x$ on every mutated model $f_i\in F$. We define LCR on a sample $x$ as follows (with respect to $F$).
\[
\varsigma(x)=\frac{|\{f_i|f_i\in F\ and\ f_i(x)\neq f(x)\}|}{|F|}
\]
, where $|S|$ is the number of elements in a set $S$. Intuitively, $\varsigma(x)$ measures how sensitive an input sample $x$ is on the mutations of a DNN model. 

Table~\ref{tb:exm} summarizes our empirical study on measuring $\varsigma(x)$ using two popular dataset, i.e., the MNIST and CIFAR10 dataset, and multiple state-of-the-art attacking methods. A total of 500 mutated models are generated using NAI operator which randomly selects some neurons and changes their activation status. The first column shows the name of the dataset. The second shows the mutation rate, i.e., the percentage of the neurons whose activation status are changed. The third shows the average LCR (with confidence interval of 99\% significance level) of 1000 normal samples randomly selected from the testing set. The remaining columns show the average LCR (with confidence interval of 99\% significance level) of 1000 adversarial samples which are generated using state-of-the-art methods. Note that column `Wrongly Labeled' are samples from the testing set which are wrongly labeled by the original DNN model. 

%for each kind of attack under 500 NAI mutated models for MNIST and CIFAR10 dataset. 
Based on the results, we can observe that at any mutation rate, the $\varsigma$ values of the adversarial samples are significantly higher than those of the normal samples.

\begin{center}
\begin{framed}
\label{obs}
\noindent \emph{$\varsigma_{adv}$ is significantly larger than $\varsigma_{nor}$.}
\end{framed}
\end{center}
%\centerline{\framebox{$\varsigma_{adv}$ is significantly larger than $\varsigma_{nor}$.}\label{obs}}
Further study on the LCR distance between normal and adversarial samples with respect to different model mutation operators is presented in Section~\ref{sec:ev}. The results are consistent.
A practical implication of the observation is that given an input sample $x$, we could potentially detect whether $x$ is likely to be normal or adversarial by checking $\varsigma(x)$. %The remaining challenge is then how to systematically  

\begin{table*}[t]
\caption{Average $\varsigma$ (shown in percentage with confidence interval of 99\% significance level) for normal samples and adversarial samples under 500 NAI mutated models.}
\centering
\begin{tabular}[h]{c|c|c|ccccccc}
\toprule
\multirow{2}{*}{Dataset} & \multirow{2}{*}{Mutation rate} & \multirow{2}{*}{Normal samples} 
&     \multicolumn{6}{c}{Adversarial samples}\\
                            & &  & Wrong labeled & FGSM     & JSMA                    &  C\&W             &Black-Box  &Deepfool  \\
\midrule
 \multirow{3}{*}{MNIST}    &0.01   &$1.28\pm0.24$     &$14.58\pm2.64$   &$47.56\pm3.56$     &$50.80\pm2.46$    &$12.07\pm1.26$  & $44.94\pm3.43$  & $37.62\pm2.83$ \\
                                      &0.03   &$3.06\pm0.44$     &$27.16\pm3.11$   &$52.12\pm3.04$     &$57.86\pm2.02$    &$21.88\pm1.38$  & $51.15\pm2.91$  & $46.61\pm2.43$ \\
                                                &0.05   &$3.88\pm0.53$     &$32.53\pm3.15$   &$54.54\pm2.80$     &$59.07\pm1.95$    &$27.73\pm1.37$  & $53.97\pm2.67$  & $50.30\pm2.24$  \\
                                                \midrule
 \multirow{3}{*}{CIFAR10}  &0.003   &$2.20\pm0.55$     & $17.95\pm1.39$  &$14.06\pm1.33$     &$28.65\pm1.30$    &$19.77\pm1.41$   &$10.36\pm1.06$  &$30.84\pm1.37$  \\
                            &0.005   &$5.05\pm0.91$     & $32.18\pm1.62$ &$27.87\pm1.71$     &$47.75\pm1.27$    &$33.95\pm1.60$   &$21.66\pm1.38$  &$47.70\pm1.23$   \\
                            
                            &0.007   &$7.28\pm1.12$     & $39.76\pm1.70$ &$36.19\pm1.81$     &$56.02\pm1.29$    &$41.22\pm1.64$   &$27.57\pm1.5$  &$54.41\pm1.21$    \\
\bottomrule
\end{tabular}
\label{tb:exm}
\end{table*}

\subsection{Explanatory Model}

\begin{figure}[t]
\centering
\includegraphics[width=.4\textwidth]{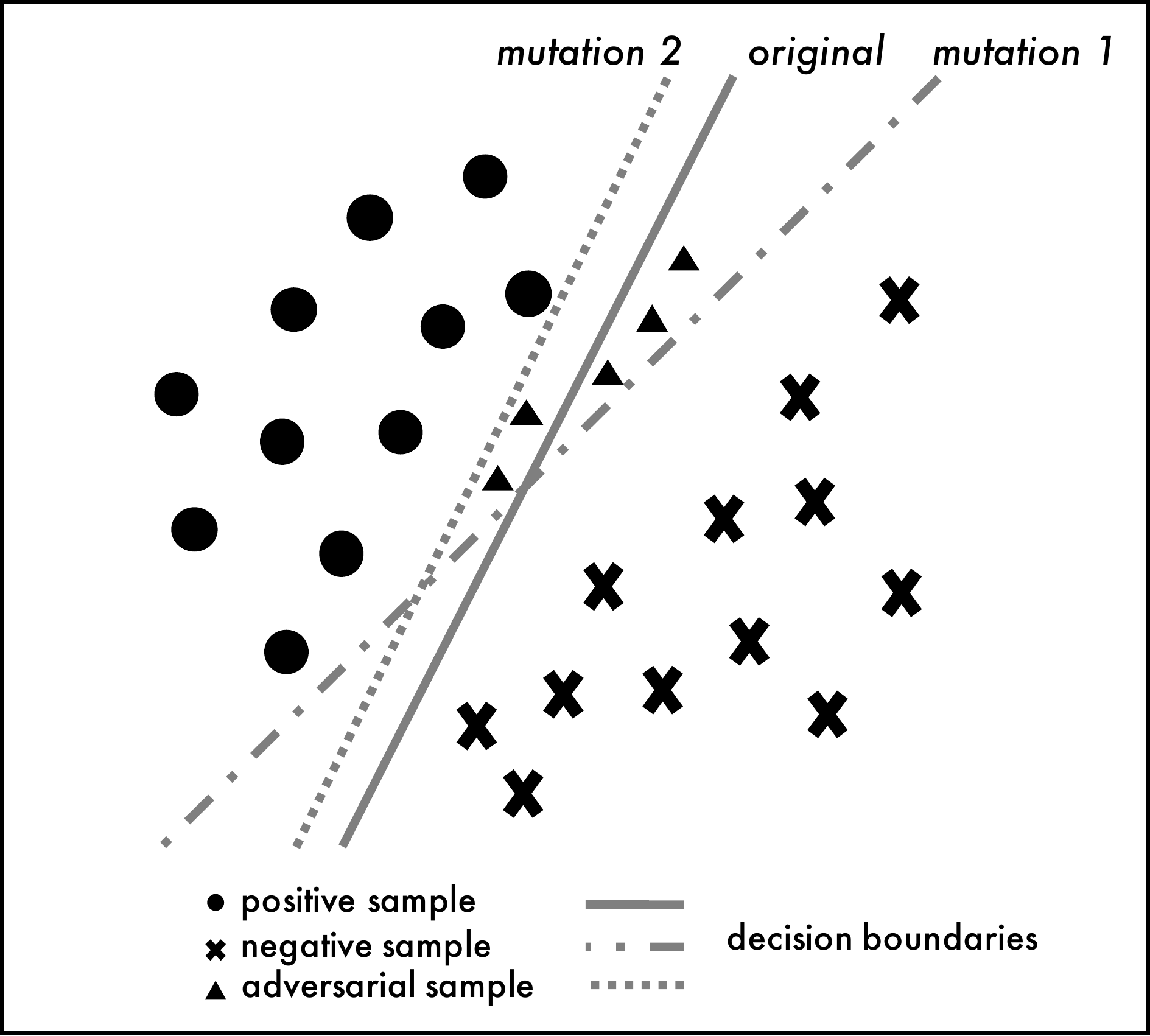}
\caption{An explanatory model of the model mutation testing effect.}
\label{fig:model}
\end{figure}

In the following, we use a simple model to explain the above observation. Recall that adversarial samples are generated in a way which tries to minimize the modification to a normal sample while is still able to cross the decision boundary. Different kinds of attacks use different approaches to achieve this goal. Our hypothesis is that most adversarial samples generated by existing methods are near the decision boundary (to minimize the modification). As a result, as we randomly mutate the model and perturb the decision boundary, adversarial samples are more likely to cross the mutated decision boundaries, i.e., if we feed an adversarial sample to a mutated model, the output label has a higher chance to change from its original label. This is illustrated visually in Figure~\ref{fig:model}.

\subsection{The Detection Algorithm}
The results shown in Table~\ref{tb:exm} suggests that we can use LCR to distinguish adversarial samples and normal samples. In the following, we present an algorithm which is designed to detect adversarial samples at runtime based on measuring the LCR of a given sample. The algorithm is based on the idea of statistical model checking~\cite{sprt,smc_survey}.

The inputs of our algorithm are a DNN model $f$, a sample $x$ and a threshold $\varsigma_{h}$ which is used to decide whether the input is adversarial. We will discuss later on how to identify $\varsigma_{h}$ systematically. The basic idea of our algorithm is to use hypothesis testing to decide the truthfulness of two mutual exclusive hypothesis.
\begin{align*}
    H_0:~~& \varsigma(x)>\varsigma_{h} \\
    H_1:~~& \varsigma(x)\le\varsigma_{h}
\end{align*}
Three (standard) additional parameters, $\alpha$, $\beta$ and $\delta$, are used to control the probability of making an error. That is, we would like to guarantee that the probability of a Type-\uppercase\expandafter{\romannumeral1} (respectively, a Type-\uppercase\expandafter{\romannumeral2}) error, which rejects $H_0$ (respectively, $H_1$) while $H_0$ (respectively, $H_1$) holds, is less or equal to $\alpha$ (respectively, $\beta$). The test needs to be relaxed with an indifferent region $(r-\delta,r+\delta)$, where neither hypothesis is rejected and the test continues to bound both types of errors~\cite{smc_survey}. In practice, the parameters (i.e., $(\alpha, \beta)$, and $\delta$) can often be decided by how much testing resources are available. In general, more resource is required for a smaller error bound.

Our detection algorithm keeps generating accurate mutated models (with an accuracy more than certain threshold on the testing data) from the original model and evaluating $\varsigma(x)$ until a stopping condition is satisfied. We remark that in practice we could generate a set of accurate mutated models before-hand and simply use them at runtime to further save detection time.
%However, such an approach is likely to be less safe if we assume that the attacker could potentially find out what the mutated models are. 

There are two main methods to decide when the testing process can be stopped, i.e., we have sufficient confidence to reject a hypothesis. One is the fixed-size sampling test (FSST), which runs a predefined number of tests. One difficulty of this approach is to find an appropriate number of tests to be performed such that the error bounds are valid. The other approach is the sequential probability ratio test (SPRT~\cite{sprt}). SPRT dynamically decides whether to reject or not a hypothesis every time after we update $\varsigma(x)$, which requires a variable number of mutated models. SPRT is usually faster than FSST as the testing process ends as soon as a conclusion is made. 

In this work, we use SPRT for the detection. The details of our SPRT-based algorithm is shown in Algorithm~\ref{alg:sprt}. The inputs of the detection algorithm include the input sample $x$, the original DNN model $f$, a mutation rate $\gamma$, and a threshold of LCR $\varsigma_h$. Besides, the detection is error bounded by $\langle\alpha,\beta\rangle$ and relaxed with an indifference region $\delta$. 
% To apply FSST, we first calculate an optimal number of mutation models needed as well as a decision threshold for the number of label change at line 3.
%\emph{We need to show the algorithm for calculating $n$ and $c$.}
% We then generate the designated number of mutated models at line 4. \emph{TODO: We need to show the algorithm for generating the mutated model with respect to certain mutation rate.} The input sample is then evaluated by all the mutated models in the while loop. After comparing the number of label changes with $c$, we make a decision whether the input sample is adversarial or not at line 10 and line 14. 
To apply SPRT, we keep generating accurate mutated models at line 5. The details of generating mutated models using the four operators in Table~\ref{tb:mu} are shown in Algorithm~\ref{alg:nai}, Algorithm~\ref{alg:gf}, Algorithm~\ref{alg:ws}, and Algorithm~\ref{alg:ns} respectively. We then evaluate whether $f_i(x)=f(x)$ at line 7.
If we observe a label change of $x$ using the mutated model $f_i$, we calculate and update the SPRT probability ratio at line 9 as:
\[
pr=\frac{p_1^{z}(1-p_1)^{n-z}}{p_0^{z}(1-p_0)^{n-z}}
\]
, with $p_1=\varsigma_h-\delta$ and $p_0=\varsigma_h+\delta$. The algorithm stops whenever a hypothesis is accepted either at line 11 or line 14. We remark that SPRT is guaranteed to terminate with probability 1~\cite{sprt}. 

We briefly introduce the NAI operator shown in Algorithm~\ref{alg:nai} as an example of the four mutation operators. We first obtain the set of $N$ unique neurons\footnote{For convolutional layer, each slide of convolutional kernel is regarded as a neuron} at line 1. Then we randomly select $\lceil N\times\gamma \rceil$ neurons ($\gamma$ is the mutation rate) for activation status inverse at line 2. Afterwards, we traverse the model $f$ layer by layer at line 3 and take those selected neurons at line 4. We then inverse the activation status of the selected neurons by multiplying their weights with -1 at line 7. 
% TODO: XXX: add one paragraph on one of the mutation operator algorithm
%On termination of Algorithm~\ref{alg:det2}, we either report the sample as a normal one or an adversarial one with an error bound.

% \begin{algorithm}[t]
% \caption{FSST-Detect(x, f, \rho, \varsigma_h, \alpha, \beta, \sigma)$}
% \label{alg:fsst}

% Let $n$ be the optimal number of needed mutations\;
% Let $c$ be the least number of mutated models $f_i$ that satisfy $f_i(x)\neq f(x)$\;
% Calculate the optimal $\langle n,c\rangle$ for $\langle\alpha,\beta,\sigma\rangle$\;
% Let $F$ be a set of $n$ randomly generated accurate mutations of $f$ using mutation rate $\pho$\;
% Let $z=0$ be the number of mutation models satisfying $f_i(x)\neq f(x)$\;
% \While {F is not empty} {
%   Take a mutation model $f_i$ from $F$\;
%   \If {$f_i(x)\neq f(x)$}{
%     $z=z+1$\;
%     \If {$z>c$}{
%       Accept the hypothesis that $\varsigma(x)>\varsigma_h$ and report the input as an adversarial input with error bounded by $\beta$\;
%       \Return \;
%     }
%   }
%   Remove $f_i$ from $F$\;
% }
% \If {$z\le c$} {
%       Accept the hypothesis that $\varsigma(x)\le\varsigma_h$ and report the input as a normal input with error bounded by $\alpha$\;
%       \Return \;
% }
% \end{algorithm}

\begin{algorithm}[t]
\caption{$\text{SPRT-Detect}(x, f, \gamma, \varsigma_h, \alpha, \beta, \delta)$}
\label{alg:sprt}

Let $stop = false$\;
Let $z=0$ be the number of mutated models $f_i$ that satisfy $f_i(x)\neq f(x)$\;
Let $n=0$ be the total number of generated mutated models so far\;
\While {!stop} {
  Apply a mutation operator to randomly generate an accurate mutation model $f_i$ of $f$ with mutation rate $\gamma$\;
  $n=n+1$\;
  \If {$f_i(x)\neq f(x)$}{
    $z=z+1$\;
    Calculate the SPRT probability ratio as $pr$\;
    \If {$pr\le \frac{\beta}{1-\alpha}$}{
      Accept the hypothesis that $\varsigma(x)>\varsigma_h$ and report the input as an adversarial sample with error bounded by $\beta$\;
      \Return \;
    }
    \If {$pr\ge \frac{1-\beta}{\alpha}$} {
      Accept the hypothesis that $\varsigma(x)\le\varsigma_h$ and report the input as a normal sample with error bounded by $\alpha$\;
      \Return \;
    }
  }
}
\end{algorithm}

\begin{algorithm}[t]
\caption{$NAI(f,\gamma)$}
\label{alg:nai}

Let $N$ be the set of unique neurons\;
Randomly select $\lceil N\times\gamma \rceil$ unique neurons\;
\For{every layer in $f$}{
    Let $Q$ be the set of selected neurons in this layer\;
    \If{$Q \neq \emptyset$}{
        \For{$q \gets Q$}{
           $q.weight = -1 \cdot q.weight$\;
        }
 }
}
\end{algorithm}

\begin{algorithm}[t]
\caption{$GF(f,\gamma)$}
\label{alg:gf}
Let $W$ be the parameters of $f$\;
Extract the parameters of $f$ layer by layer\;
Let $N$ be the total number of parameters of $f$\;
Randomly select $\lceil N\times\gamma \rceil$ parameters to fuzz\;
\For{every layer in $f$}{
    Let $W[i]$ be the parameters of this layer\;
    Find all the selected parameters $P$ in $W[i]$\;
    \If{$P\neq \emptyset$}{
    Let $\mu= Avg(W[i])$\;
    Let $\sigma= Std(W[i])$\;
    \For{every parameter in $P$}{
        Randomly assign the parameter according to $\mathcal{N}(\mu,\sigma^{2})$\;
    }
    }
}
\end{algorithm}

\begin{algorithm}[t]
\caption{$WS(f,\gamma)$}
\label{alg:ws}

Let $N$ be the set of unique neurons\;
Randomly select $\lceil N\times\gamma \rceil$ unique neurons to shuffle \;
\For{every layer in $f$}{
    Let $Q$ be the set of selected neurons in this layer\;
    \If{$Q \neq \emptyset$}{
        \For{$q \gets Q$}{
           $q.weight =$ Shuffle$(q.weight)$\;
        }
 }
}
\end{algorithm}

\begin{algorithm}[t]
\caption{$NS(f,\gamma)$}
\label{alg:ns}

\For{every layer in $f$}{
    Let $N$ be the number of unique neurons in this layer\;
    Randomly select $\lceil N\times\gamma \rceil$ unique neurons\;
    Let $Q$ be the set of selected neurons\;
    Randomly switch the weights of neurons in $Q$\;
 }
\end{algorithm}

% \subsection*{Overall Algorithm}

% \input{experiment.tex}

% \emph{TODO: We need to briefly discuss the complexity of both algorithms. FSST is always terminating?}

\section{Implementation and Evaluation}\label{sec:ev}
We have implemented our approach in a self-contained toolkit which is available online~\cite{tool}. It is implemented in Python with about 5k lines of code. In the following, we evaluate the accuracy and efficiency of our approach through multiple experiments. 

\subsection{Experiment Settings} 
\paragraph{Datasets and Models} We adopt two popular image datasets for our evaluation: MNIST and CIFAR10. Each dataset has 60000/50000 images for training and 10000/10000 images for testing. The target models for MNIST and CIFAR10 are LeNet~\cite{lenet} and GooglLeNet~\cite{googlenet} respectively. The accuracy of our trained models on training and testing dataset are 98.5\%/98.3\% for MNIST and 99.7\%/90.5\% for CIFAR10 respectively, which both achieve state-of-the-art performance.
% The two targeted models are trained on the complete training data and the accuracy is reported on testing data, say $98.3\%$ for LeNet on MNIST and $90.5\%$ for GoogLeNet on CIFAR10.

% \begin{table}[t]
% \centering
% \caption{Datasets for evaluation}
% \label{tb:ds}
% \begin{tabular}{@{}lllll@{}}
% \toprule
% Dataset                 & \#Training & \#Testing & \#Target classes \\ \midrule
% MNIST                   & 60000      & 10000     & 10               \\
% CIFAR10                 & 60000      & 10000     & 10               \\ \bottomrule
% \end{tabular}
% \end{table}

\paragraph{Mutated models generation} We employ the four mutation operators shown in Table~\ref{tb:mu} to generate mutated models.
% Notice that the convolutional layer is quite different from the linear layer where a neuron is clearly defined, we regard each slide of convolutional kernel as a neuron. As a result, 
In total, we have $236$ neurons for the MNIST model and $7914$ neurons for the CIFAR10 model. For each mutation operator, we generate three groups of mutation models from the original trained model using different mutation rate to see its effect. The mutation rate we use for the MNIST model is $\{0.01,0.03,0.05\}$ and $\{0.003,0.005,0.007\}$ for the CIFAR10 model (since there are more neurons). Note that some mutation models may have significantly worse performance, so not all mutated models are valid. In our experiment, we only keep those mutation models whose accuracy on the testing dataset is not lower than $90\%$ of that of its seed model. For each mutation rate, we generate 500 such accurate mutated models for our experiments. \\

\paragraph{Adversarial samples generation}
We test our detection algorithm against four state-of-the-art attacks in Clverhans~\cite{papernot2017cleverhans} and Deepfool~\cite{DeepFool} (detailed in Section~\ref{sec:back}). For each kind of attack, we generate a set of adversarial samples for evaluation. The parameters for each kind of attack to generate the adversarial samples are summarized as follows.
\begin{itemize}

\item \textbf{FGSM}: There is only one parameter to control the scale of perturbation. We set it as 0.35 for MNIST and 0.03 for CIAFR10 according to the original paper.

\item \textbf{JSMA}: There is only one parameter to control the maximum distortion. We set it as  $12\%$ for both datasets, which is slightly smaller than the original paper.

\item \textbf{C\&W}: There are three types of attacks proposed in~\cite{CW}: $L_0$, $L_2$ and $L_{\infty}$. We adopt $L_2$ attack according to the author's recommendation. We also set the scale coefficient to be 0.6 for both datasets. We set the iteration number to be 10000 for MNIST and 1000 for CIFAR10 according to the original paper.

\item \textbf{Deepfool}: We set the maximum number of iterations to be 50 and the termination criterion (to prevent vanishing updates) to be 0.02 for both datasets, which is a default setting in the original paper.

\item \textbf{Black-Box}: The key setting of the Black-Box attack is to train a substitute model of the target model. The substitute model for MNIST is the first model defined in Appedix A of~\cite{bb}. For CIFAR10, we use the LeNet~\cite{lenet} as the surrogate model. Afterwards, the attack algorithm we used for the surrogate model is FGSM.
\end{itemize}

\begin{table}[t]
\centering
\caption{Number of samples in each group.}
\label{tb:adv-data}
\begin{tabular}{c|c|cccccc}
\toprule
Dataset & Attack          & Samples \\ \midrule
\multirow{7}{*}{MNIST}   & Normal          &     1000    \\
        & Wrongly-labeled &  171    \\
        & FGSM            &  1000       \\
        & JSMA            &  1000        \\
        & BB              &  1000       \\
        & C\&W            &  743       \\
        & Deepfool        &  1000        \\ \midrule
\multirow{7}{*}{CIFAR10} & Normal          &    1000     \\
        & Wrongly-labeled &     951    \\
        & FGSM            &     1000    \\
        & JSMA            &     1000    \\
        & BB              &     1000    \\
        & C\&W            &     1000    \\
        & Deepfool        &     1000    \\ \bottomrule
\end{tabular}
\end{table}

% \begin{table*}[t]
% \centering
% \caption{Number of samples for evaluation in each group.}
% \label{tb:adv-data}
% \begin{tabular}{c|c|cccccc}
% \toprule
% \multirow{2}{*}{Dataset} & \multirow{2}{*}{Normal Samples} & \multicolumn{6}{c}{Adversarial Samples}\\
%     & & FGSM & JSAM & Black-Box & C\&W & deepfool  & Wrongly Labeled\\
% \midrule
% MNSIT   & 2000 &2000 & 2000   &? &? &2000 & 171\\
% CIFAR10 & 2000 &2000 & 2000   &? &1079 &2000 & 951 \\
% \bottomrule
% \end{tabular}\\
% \end{table*}

For each attack, we make 1000 attempts to generate adversarial samples. Notice that not all attempts are successful and as a result we manage to generate no more than 1000 adversarial samples for each attack. Further recall that according to our definition, those samples in the testing dataset which are wrongly labeled by the trained DNN are also adversarial samples. Thus, in addition to the adversarial samples generated from the attacking methods, we attempt to randomly select 1000 samples from the testing dataset which are wrongly classified by the target model as well. Table~\ref{tb:adv-data} summarizes the number of normal samples and valid adversarial samples for each kind of attack used for the experiments.

\subsection{Evaluation Metrics}

\paragraph{Distance of label change rate} We use $d_{lcr}=\varsigma_{adv}/\varsigma_{nor}$ where $\varsigma_{adv}$ (and $\varsigma_{nor}$) is the average LCR of adversarial samples (and normal samples) to measure the distance between the LCR of adversarial samples and normal samples. The larger the value is, the more significant is the difference.

\paragraph{Receiver characteristics operator} Since our detection algorithm works based on a threshold LCR $\varsigma_h$, we first adopt receiver characteristic operator (ROC) curve to see how good our proposed feature, i.e., LCR under model mutation, is to distinguish adversarial and normal samples~\cite{roc,feinman2017detecting}. 
% \emph{For the SE community, you need to explain a bit more on what ROC is.} 
The ROC curve plots the true positive rate ($tpr$) against false positive rate ($fpr$) for every possible threshold for the classification. From the ROC curve, we could further calculate the area under the ROC curve (AUROC) to characterize how well the feature performs. A perfect classifier (when all the possible thresholds yield true positive rate 1 and false positive rate 0 for distinguishing normal and adversarial samples) will have AUROC 1. The closer is AUROC to 1, the better is the feature.

\paragraph{Accuracy of detection} The accuracy of the detection is defined in a standard way as follows. Given a set of images $X$ (labeled as normal or adversarial), what is the percentage that our algorithm correctly classifies it as normal or adversarial? Notice that the accuracy of detecting adversarial samples is equivalent to $tpr$ and the accuracy of detecting normal samples is equivalent to $1-fpr$. The higher the accuracy, the better is our detection algorithm.
%The higher the accuracy is, the better is our detection algorithm. We also use true positive rate (higher the better) and false positive rate (lower the better) to report the sensitivity and specificity of the detection respectively.

\subsection{Research Questions}

\noindent \emph{RQ1: Is there a significant difference between the LCR of adversarial samples and normal samples under different model mutations?} To answer the question, we calculate the average LCR of the set of normal samples and the set of adversarial samples generated as described above with a set of mutated models using different mutation operators. A set of 500 mutants are generated for each mutation operator (note that mutation rate 0.003 is too low for NS to generate mutated models for CIFAR10 model and thus omitted). According to the detailed results summarized in Tabel~\ref{tb:exm} and~\ref{tb:lcr}, we have the following answer.

\begin{framed}
\noindent \emph{Answer to RQ1: Adversarial samples have significantly higher LCR under model mutation than normal samples.
}\end{framed}

\begin{table*}[t]
\footnotesize
\centering
\caption{Label change rate (confidence interval with 99\% significance level) for each group of samples under model mutation testing with different mutation operators (NAI result is shown previously in Table~\ref{tb:exm}). The results are shown in percentage.}
\label{tb:lcr}
\begin{adjustbox}{width=\textwidth}
\begin{tabular}[h]{c|c|c|c|cccccc}
\toprule
\multirow{2}{*}{Mutation operator} & \multirow{2}{*}{Dataset} & \multirow{2}{*}{Mutation rate} & \multirow{2}{*}{Normal samples} 
&   &  \multicolumn{5}{c}{Adversarial samples}\\
                           & & &  & Wrong labeled & FGSM     & JSMA                    &  C\&W             &Black-Box  &Deepfool  \\
\midrule
% \multirow{6}{*}{NAI} & \multirow{3}{*}{MNIST}    &0.01   &$1.28\pm0.24$     &$14.58\pm2.64$   &$47.56\pm3.56$     &$50.80\pm2.46$    &$12.07\pm1.26$  & $44.94\pm3.43$  & $37.62\pm2.83$ \\
%                       &                          &0.03   &$3.06\pm0.44$     &$27.16\pm3.11$   &$52.12\pm3.04$     &$57.86\pm2.02$    &$21.88\pm1.38$  & $51.15\pm2.91$  & $46.61\pm2.43$ \\
%                       &                         &0.05   &$3.88\pm0.53$     &$32.53\pm3.15$   &$54.54\pm2.80$     &$59.07\pm1.95$    &$27.73\pm1.37$  & $53.97\pm2.67$  & $50.30\pm2.24$  \\
% & \multirow{3}{*}{CIFAR10}  &0.001   &$2.20\pm0.55$     & $17.95\pm1.39$  &$14.06\pm1.33$     &$28.65\pm1.30$    &$19.77\pm1.41$   &$10.36\pm1.06$  &$30.84\pm1.37$  \\
%         &                    &0.003   &$5.05\pm0.91$     & $32.18\pm1.62$ &$27.87\pm1.71$     &$47.75\pm1.27$    &$33.95\pm1.60$   &$21.66\pm1.38$  &$47.70\pm1.23$   \\
                            
%         &                    &0.005   &$7.28\pm1.12$     & $39.76\pm1.70$ &$36.19\pm1.81$     &$56.02\pm1.29$    &$41.22\pm1.64$   &$27.57\pm1.5$  &$54.41\pm1.21$    \\
% \midrule
\multirow{6}{*}{NS} & \multirow{3}{*}{MNIST}    &0.01   &$0.12\pm0.07$     & $3.78\pm0.94$
&$44.67\pm3.92$     &$36.03\pm3.24$    &$3.42\pm0.79$  &$40.06\pm3.82$   &$26.09\pm3.16$  \\

                                     &          &0.03   &$0.37\pm0.19$     & $10.78\pm2.30$

&$46.32\pm3.71$     &$47.45\pm2.61$    &$8.93\pm1.16$  &$43.05\pm3.59$   &$34.20\pm2.92$  \\
                                     &          &0.05   &$0.89\pm0.35$     & $19.30\pm3.18$
&$48.91\pm3.41$     &$56.51\pm2.11$    &$15.87\pm1.53$  &$46.94\pm3.29$   &$42.69\pm2.65$   \\

& \multirow{3}{*}{CIFAR10}   &0.003   & -    & -  & -     & -     & -  & - & - \\
 &                           &0.005  & $0.02\pm0.03$    &$0.3\pm0.15$   &$0.3\pm0.16$    &$0.46\pm0.16$    &$0.37\pm0.18$   & $0.08\pm0.05$ &$0.86\pm0.24$   \\
                            
  &                          &0.007   & $0.94\pm0.4$     &$10.12\pm1.19$   &$7.16\pm1.06$     &$16.07\pm1.21$    &$11.04\pm1.19$   &$4.61\pm0.8$  &$19.05\pm1.37$   \\

\midrule
\multirow{6}{*}{WS} & \multirow{3}{*}{MNIST}    &0.01   &$0.93\pm0.18$     &$9.83\pm2.33$
& $46.04\pm3.73$    &  $46.96\pm2.67$  & $7.98\pm1.15$ & $42.42\pm3.62$  &$33.41\pm2.97$  \\
                     &                          &0.03   &$3.03\pm0.35$     &$21.84\pm3.11$
& $49.83\pm3.26$    & $56.01\pm2.10$   &$17.01\pm1.38$  & $47.98\pm3.14$  &$43.07\pm2.60$  \\
                     &                          &0.05   &$3.83\pm0.42$     &$26.96\pm3.26$
& $51.46\pm3.06$    &  $57.56\pm2.00$  &$21.03\pm1.40$  & $50.20\pm2.94$  &$46.37\pm2.46$   \\
& \multirow{3}{*}{CIFAR10}  &0.003   &$0.79\pm0.35$      &$9.04\pm1.17$    &$6.43\pm1.05$     &$14.85\pm1.27$    &$10.01\pm1.18$   &$9.11\pm 0.74$  &$18.78\pm1.46$ \\
   &                         &0.005   &$2.01\pm0.55$     & $17.0\pm1.53$    &$12.88\pm0.145$     &$29.42\pm1.55$    &$18.42\pm1.55$  &$8.49\pm1.06$   &$32.63\pm1.63$  \\                   
   &                         &0.007   &$2.69\pm0.65$     &$21.6\pm1.67$    &$17.21\pm1.67$     &$37.69\pm1.63$    &$23.40\pm1.69$   &$11.15\pm1.22$  &$40.03\pm1.63$   \\
\midrule
\multirow{6}{*}{GF} & \multirow{3}{*}{MNIST}    &0.01   &$0.57\pm0.30$     &$16.75\pm3.33$   
&$47.87\pm3.54$     & $56.39\pm2.14$   & $14.27\pm1.56$ & $45.56\pm3.41$  &$41.07\pm2.76$  \\
                     &                          &0.03   &$1.39\pm0.46$     &$27.00\pm3.40$   
&$51.87\pm3.10$     & $60.64\pm1.85$   & $22.10\pm1.64$ & $50.59\pm2.97$  &$48.06\pm2.41$  \\
                     &                          &0.05   &$2.49\pm0.59$     &$33.28\pm3.28$   
&$55.02\pm2.77$     & $62.36\pm1.74$   & $25.87\pm1.55$ & $53.38\pm2.68$  &$51.60\pm2.19$   \\

& \multirow{3}{*}{CIFAR10}  &0.003   &$1.42\pm0.51$     &$15.36\pm1.52$   &$11.42\pm1.42$     &$26.52\pm1.53$    &$17.0\pm1.51$   &$8.05\pm1.10$  &$31.36\pm1.68$  \\
  &                          &0.005   &$2.89\pm0.75$     &$25.31\pm1.75$   &$20.71\pm1.79$     &$41.69\pm1.54$    &$26.59\pm1.75$   &$13.75\pm1.34$  &$45.8\pm1.57$   \\
                            
   &                         &0.007   &$4.09\pm0.91$     &$31.97\pm1.86$   &$27.69\pm1.97$     &$50.07\pm1.52$    &$32.94\pm1.82$   &$18.29\pm1.48$  &$53.67\pm1.51$   \\
\bottomrule
\end{tabular}
\end{adjustbox}
\end{table*}

\begin{figure*}[t]
\begin{subfigure}[b]{0.25\textwidth}
   \centering 
   \includegraphics[height=1in]{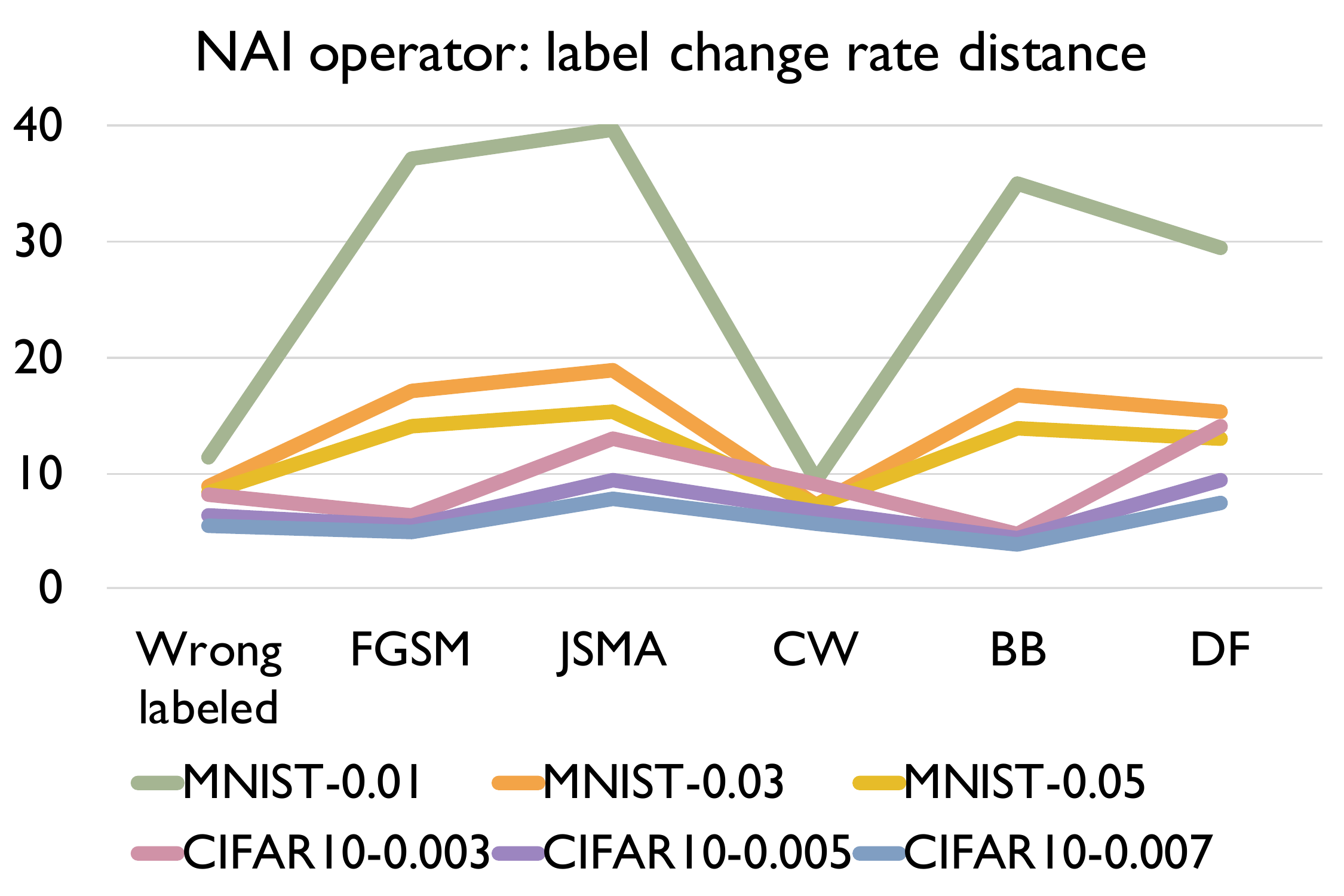}
   % \caption{NASA Logo} 
   \label{fig:dis:nai}
\end{subfigure}% 
\begin{subfigure}[b]{0.25\textwidth}
   \centering 
   \includegraphics[height=1in]{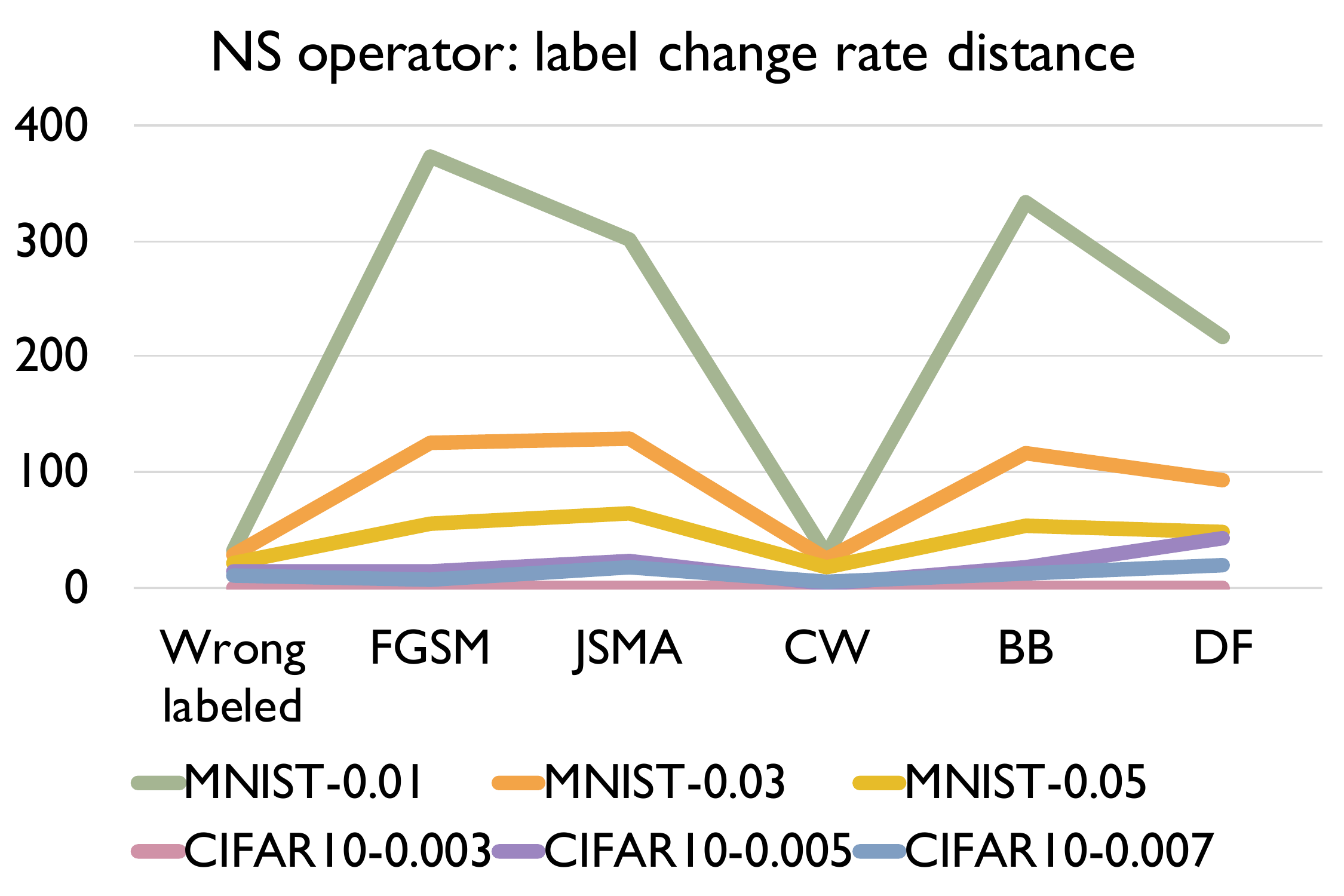}
   % \caption{Orion Logo} 
   \label{fig:dis:ns}
\end{subfigure}%
\begin{subfigure}[b]{0.25\textwidth}
   \centering 
   \includegraphics[height=1in]{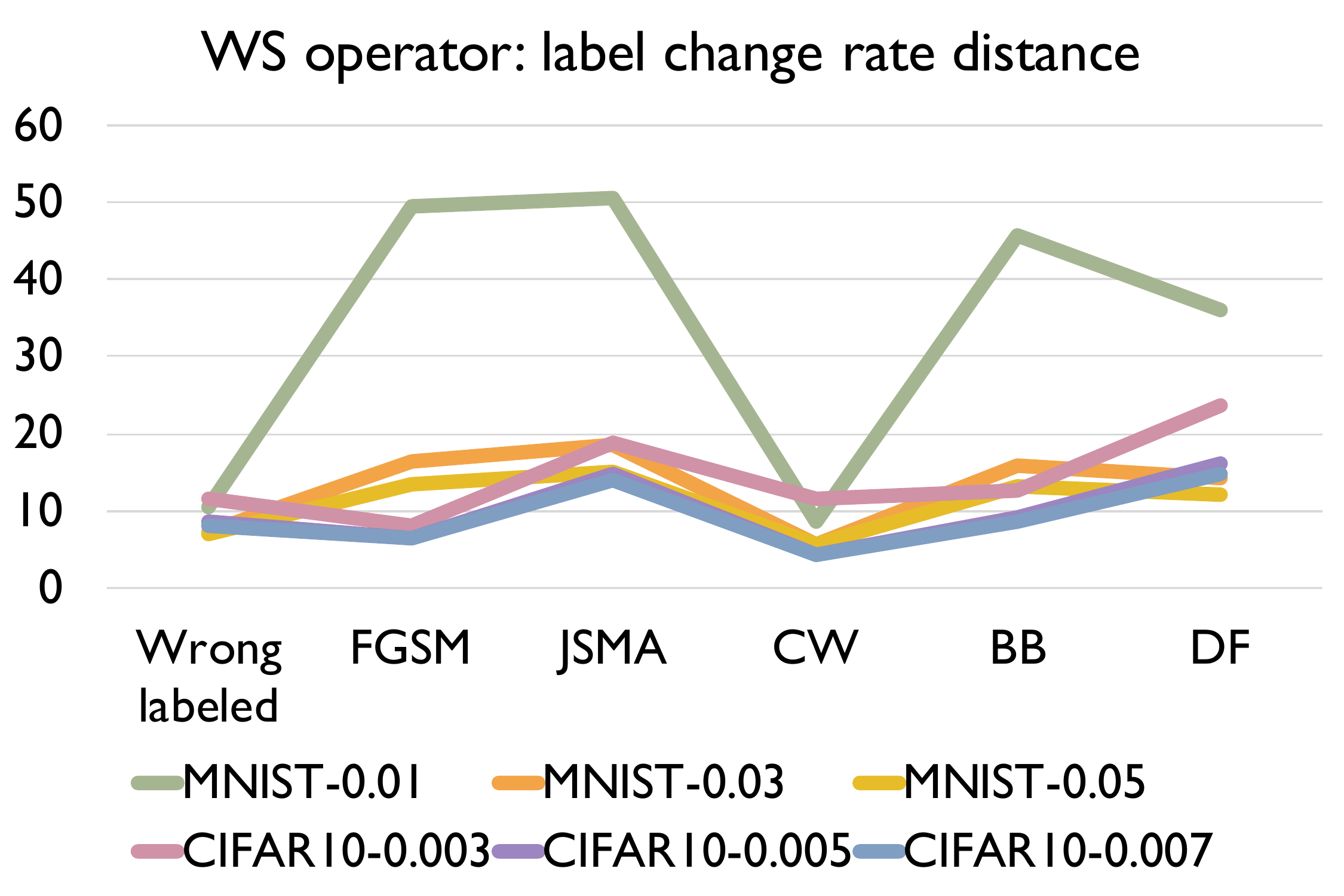}
   % \caption{NASA Logo} 
   \label{fig:dis:ws}
\end{subfigure}% 
\begin{subfigure}[b]{0.25\textwidth}
   \centering 
   \includegraphics[height=1in]{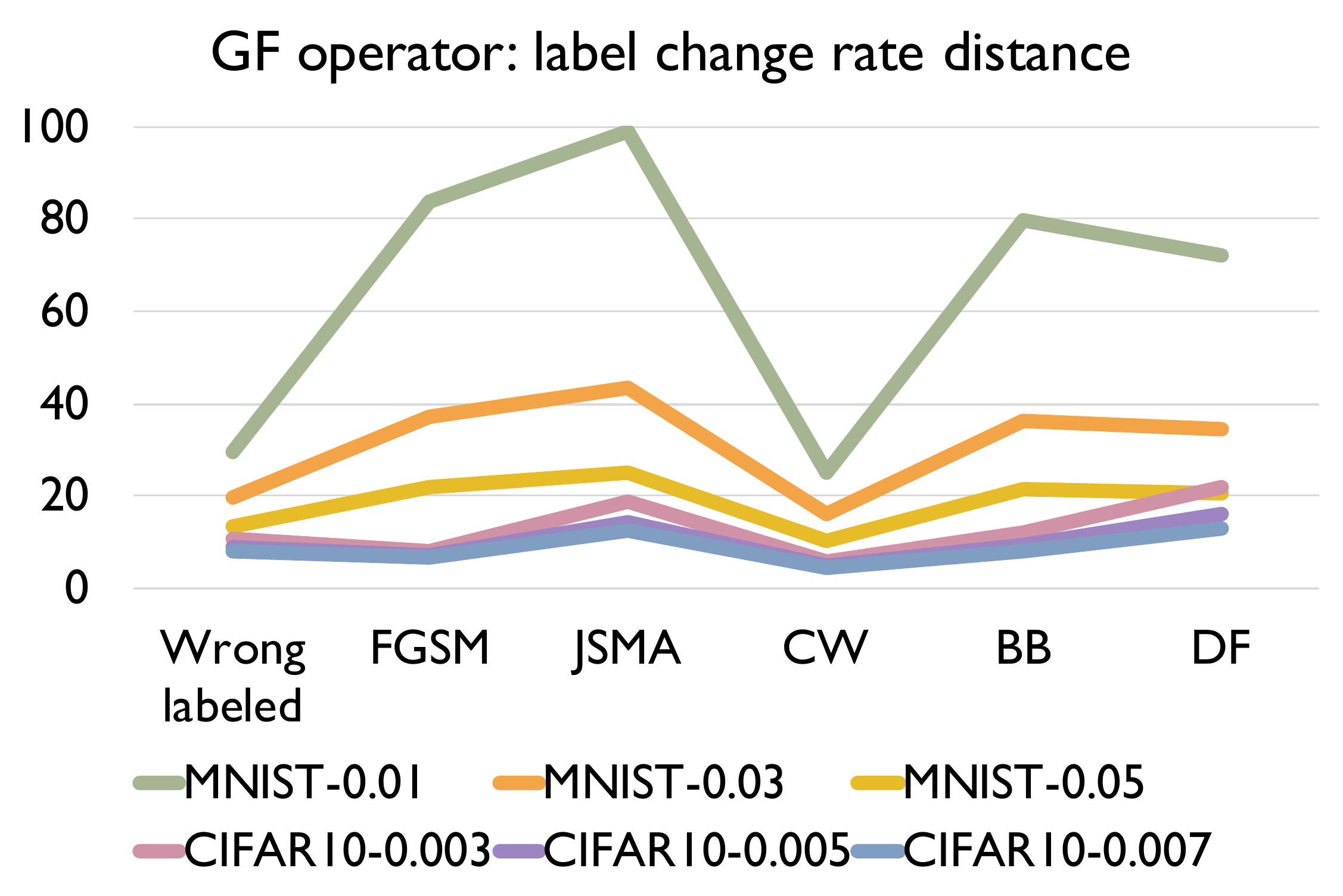}
   % \caption{Orion Logo} 
   \label{fig:dis:gf}
\end{subfigure}%
   \caption{LCR distance between normal samples and adversarial samples using different mutation operators}
   \label{fig:dis}
\end{figure*}

In addition, we have the following observations.
\begin{itemize}
\item Adversarial samples generated from every kind of attack have significantly larger LCR than normal samples under a set of mutated models under any mutation rate, and different kind of attack have different LCR. We can see that the LCR of normal samples are very low (i.e., comparable to the testing error) and that of adversarial samples are much higher. Figure~\ref{fig:dis}
shows the distance between LCR of adversarial samples and normal samples for different mutation operators. We can see that the distance is mostly larger than 10 and can be up to 375, which well supports our answer to RQ1. We can also observe that adversarial samples generated by FGSM/JSMA/Deepfool/Black-box have relatively higher LCR distance than those generated by CW and those wrong-labeled samples in the original dataset. In general, our detection algorithm is able to detect attacks with larger distance faster and better.

\item As we increase the model mutation rate, the LCR of both normal samples and adversarial samples increase (as expected) and the distance between them decreases. We can observe from Table~\ref{tb:lcr} that the LCR increases with an increasing model mutation rate in all cases. From Figure~\ref{fig:dis}, we see that a smaller model mutation rate like 0.01 for MNIST and 0.003 for CIFAR10 have the largest LCR distance. This is probably because as we increase the mutation rate, normal samples are more sensitive in terms of the change of LCR since it is a much smaller number.

\item Like adversarial samples generated by different attacking methods, wrongly labeled samples also have significantly larger LCR than normal samples. This suggests that wrongly labeled samples are also sensitive to the change of decision boundaries from model mutations as adversarial samples. They are the same as the adversarial samples which are near to the decision boundary and thus can be potentially detected.
% This also suggests that we should prefer a small model mutation rate for a larger distance to better distinguish adversarial sample from normal samples.
\end{itemize}

\noindent \emph{RQ2: How good is the LCR under model mutation as an indicator for the detection of adversarial samples?} To answer the question, we further investigate the ROC curve using LCR as the indicator of classifying an input sample as normal or adversarial. We compare our proposed feature, i.e., LCR under model mutations with two baseline approaches. The first baseline (referred as baseline 1) is a combination of density estimate and model uncertainty estimate as joint features~\cite{feinman2017detecting}. The second baseline (referred as baseline 2) is based on the label change rate of imposing random perturbations on the input sample~\cite{detection_ours}.

Table~\ref{tb:roc} presents the AUROC results under different model mutation operators. We compare our results with two baselines introduced above. The best AUROC results among the three approaches are in bold. We could observe that our proposed feature beats both baselines in over half the cases (excluding Deepfool which we do not have any reported baseline results), while baseline 1 and baseline 2 only win 1 and 3 cases respectively. We could also observe that the AUROC results are mostly very close to 1 (a perfect classifier), i.e., usually larger than 0.9, which suggests that we could achieve high accuracy using the proposed feature to distinguish adversarial samples. We thus have the following answer to RQ2.

\begin{framed}
\noindent \emph{Answer to RQ2: LCR under model mutation could outperform current baselines to detect adversarial samples.}
\end{framed}

\begin{table}[t]\centering
\caption{AUROC results. BL means `baseline'.}
\begin{adjustbox}{width=.49\textwidth}

\begin{tabular}{@{}c|c|cc|cccc@{}}
\toprule
Dataset & Attack & BL 1 & BL 2 & NAI    & GF     & NS     & WS     \\ \midrule
\multirow{6}{*}{MNIST}   & FGSM   & 0.9057     & \textbf{0.9767}    & 0.9744 & 0.9747 & 0.9554 & 0.9648 \\
        & JSMA   & 0.9813     & 0.9946    & 0.9965 & \textbf{0.9975} & 0.9975 & 0.9969 \\
        & CW     & \textbf{0.9794}     & 0.9394    & 0.9576 & 0.9521 & 0.909  & 0.9225 \\
        & BB     & --           & 0.9403    & \textbf{0.9789} & 0.9763 & 0.9631 & 0.9725 \\
        & DF     & --           & --          & 0.9881 & 0.\textbf{9889} & 0.9853 & 0.9864 \\
        & WL     & --           & 0.9696    & 0.9689 & \textbf{0.9727} & 0.9612 & 0.9692 \\
        \midrule
\multirow{6}{*}{CIFAR10}   & FGSM   & 0.7223     & \textbf{0.9099}    & 0.8956 & 0.8779 & 0.7559 & 0.8458 \\
        & JSMA   & 0.9152     & 0.8753    & 0.9733 & \textbf{0.9737} & 0.9355 & 0.9729 \\
        & CW     & 0.9217     & 0.8385    & \textbf{0.926}  & 0.9205 & 0.8464 & 0.8994 \\
        & BB     & --           & \textbf{0.9251}    & 0.874  & 0.8371 & 0.7068 & 0.8702 \\
        & DF     & --           & --          & 0.974  & \textbf{0.9786} & 0.9549 & 0.9753 \\
        & WL     & --           & 0.9148    & \textbf{0.9185} & 0.9146 & 0.8331 & 0.876  \\ \bottomrule
\end{tabular}
\end{adjustbox}
\label{tb:roc}
\end{table}

\begin{figure*}[t]
\begin{subfigure}[b]{.25\textwidth}
   \centering 
   \label{fig:acc:gf}%
   \includegraphics[height=1in]{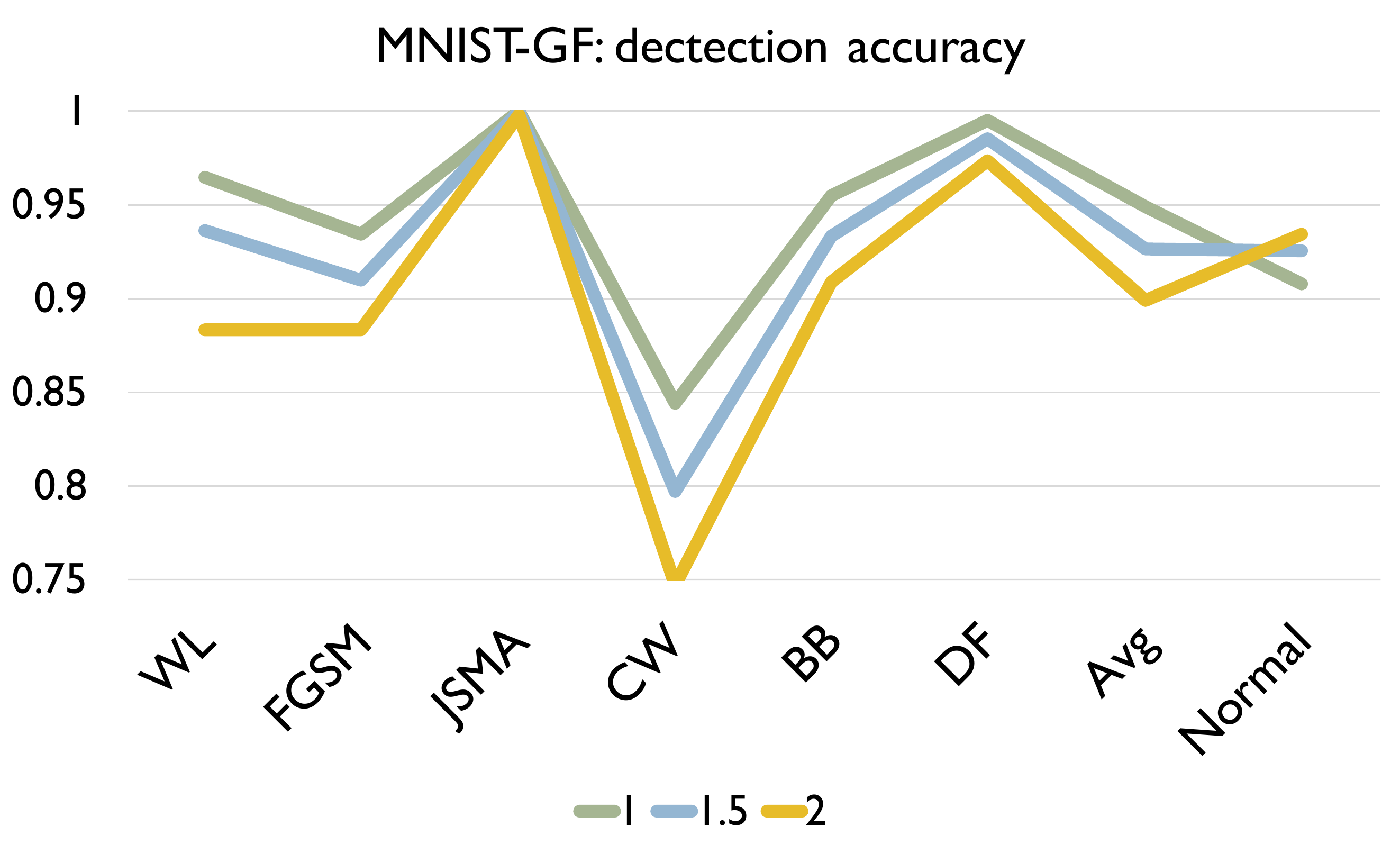}%
\end{subfigure}% 
\begin{subfigure}[b]{.25\textwidth}
   \centering 
   \label{fig:mu:gf}%
   \includegraphics[height=1in]{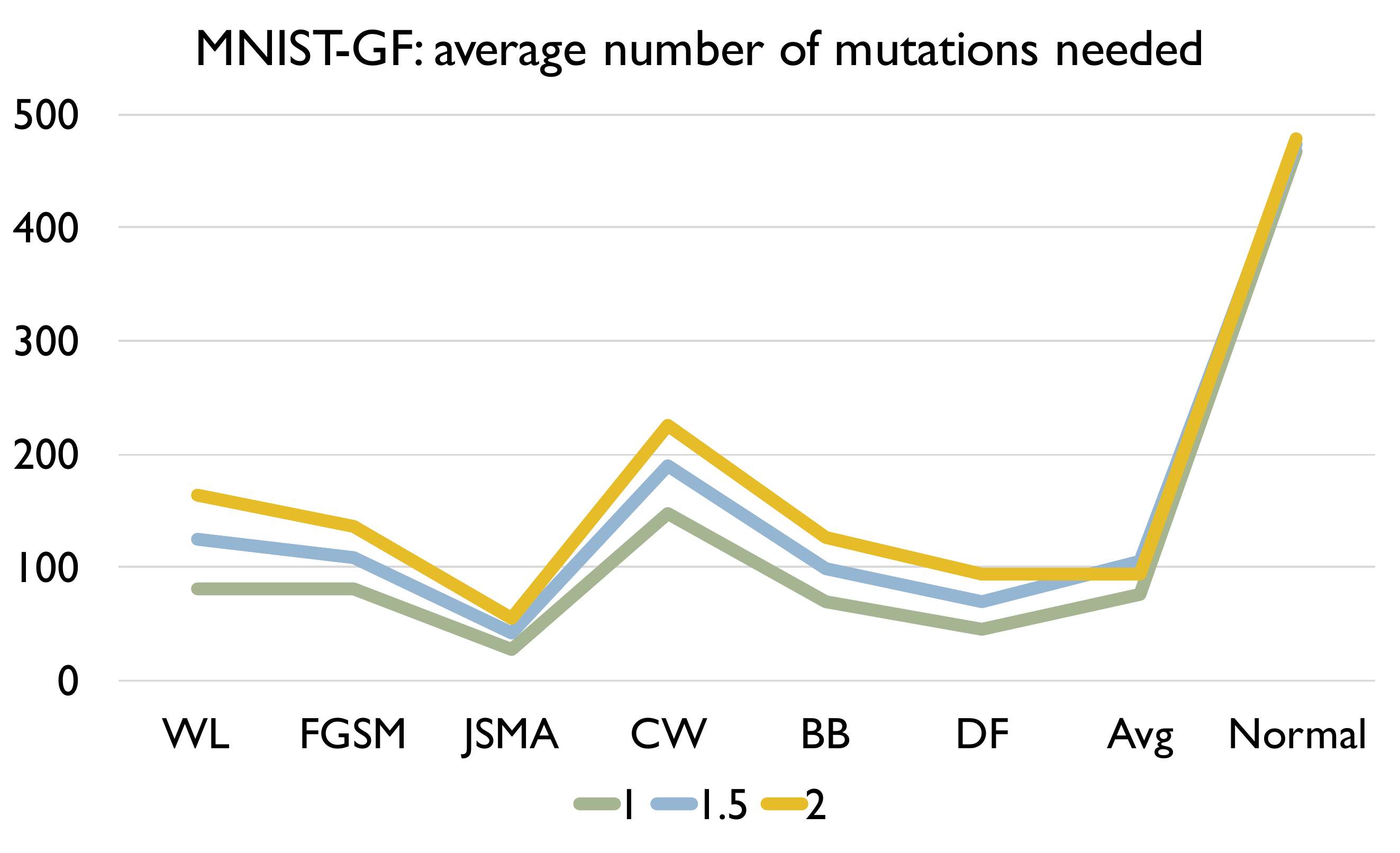}%
\end{subfigure}%
\begin{subfigure}[b]{.25\textwidth}
   \centering 
   \label{fig:acc:nai}%
   \includegraphics[height=1in]{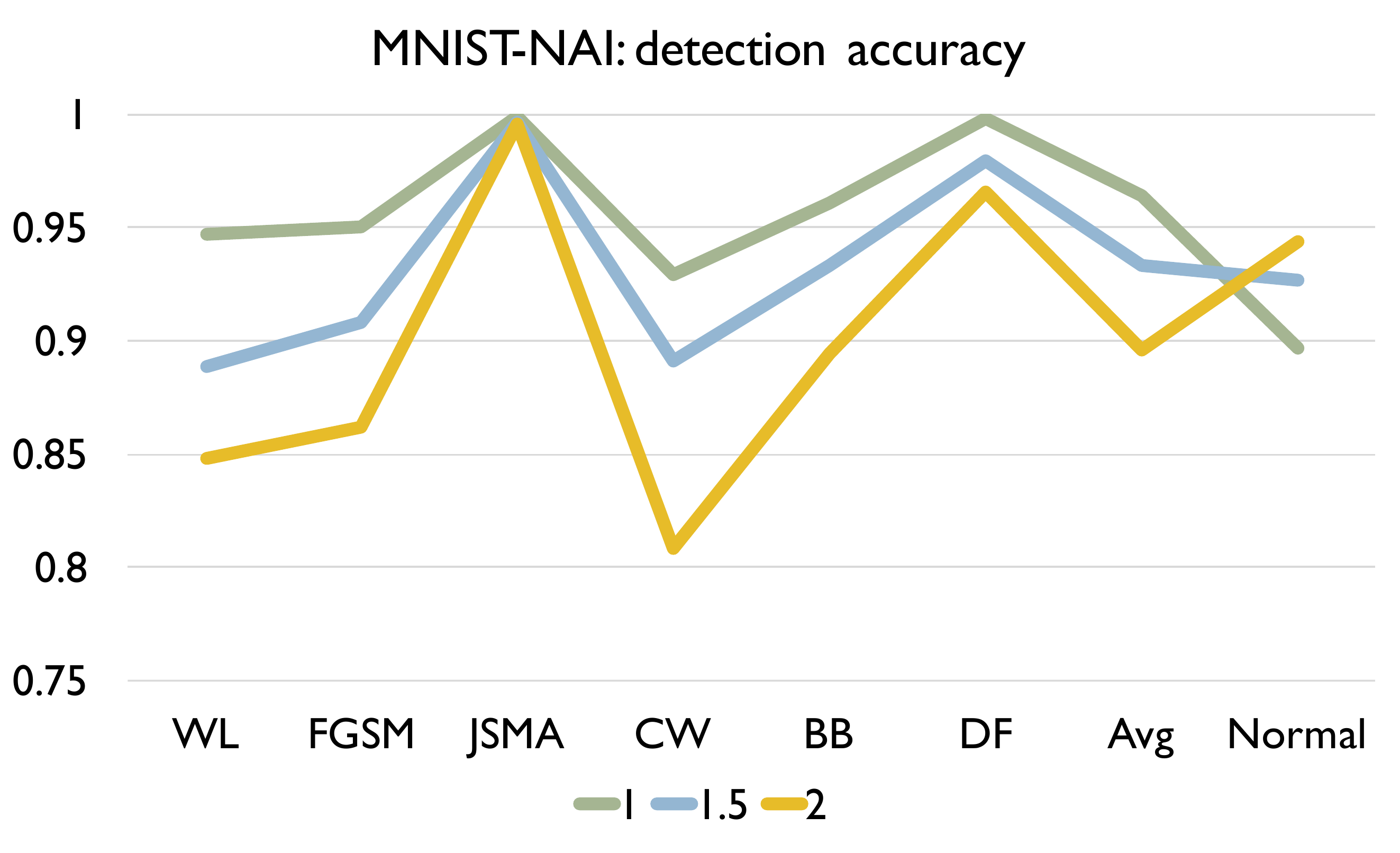}
\end{subfigure}% 
\begin{subfigure}[b]{.25\textwidth}
   \centering 
   \label{fig:mu:nai}%
   \includegraphics[height=1in]{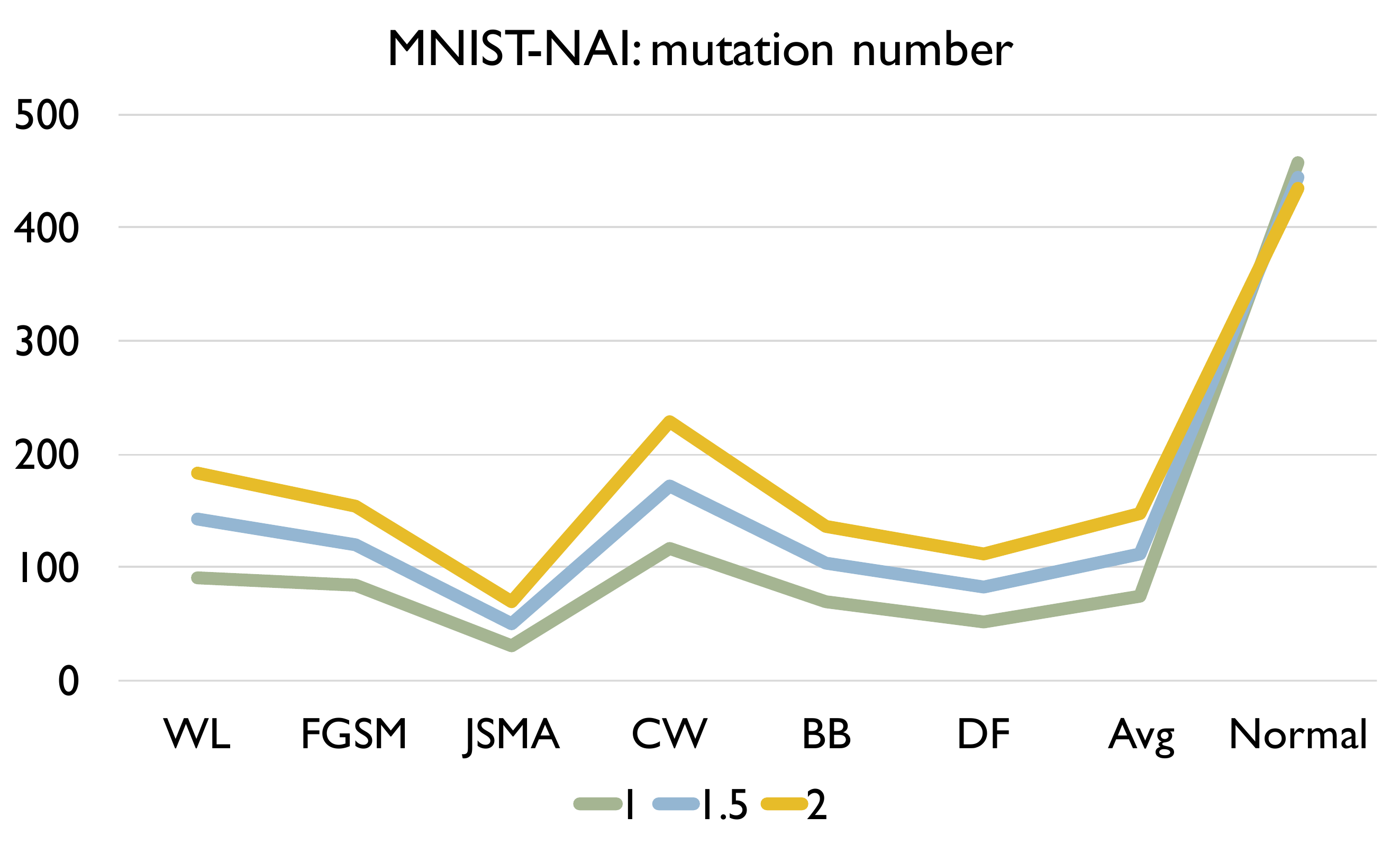}
\end{subfigure}%
\\
\begin{subfigure}[b]{.25\textwidth}
   \centering 
   \label{fig:acc:ns}%
   \includegraphics[height=1in]{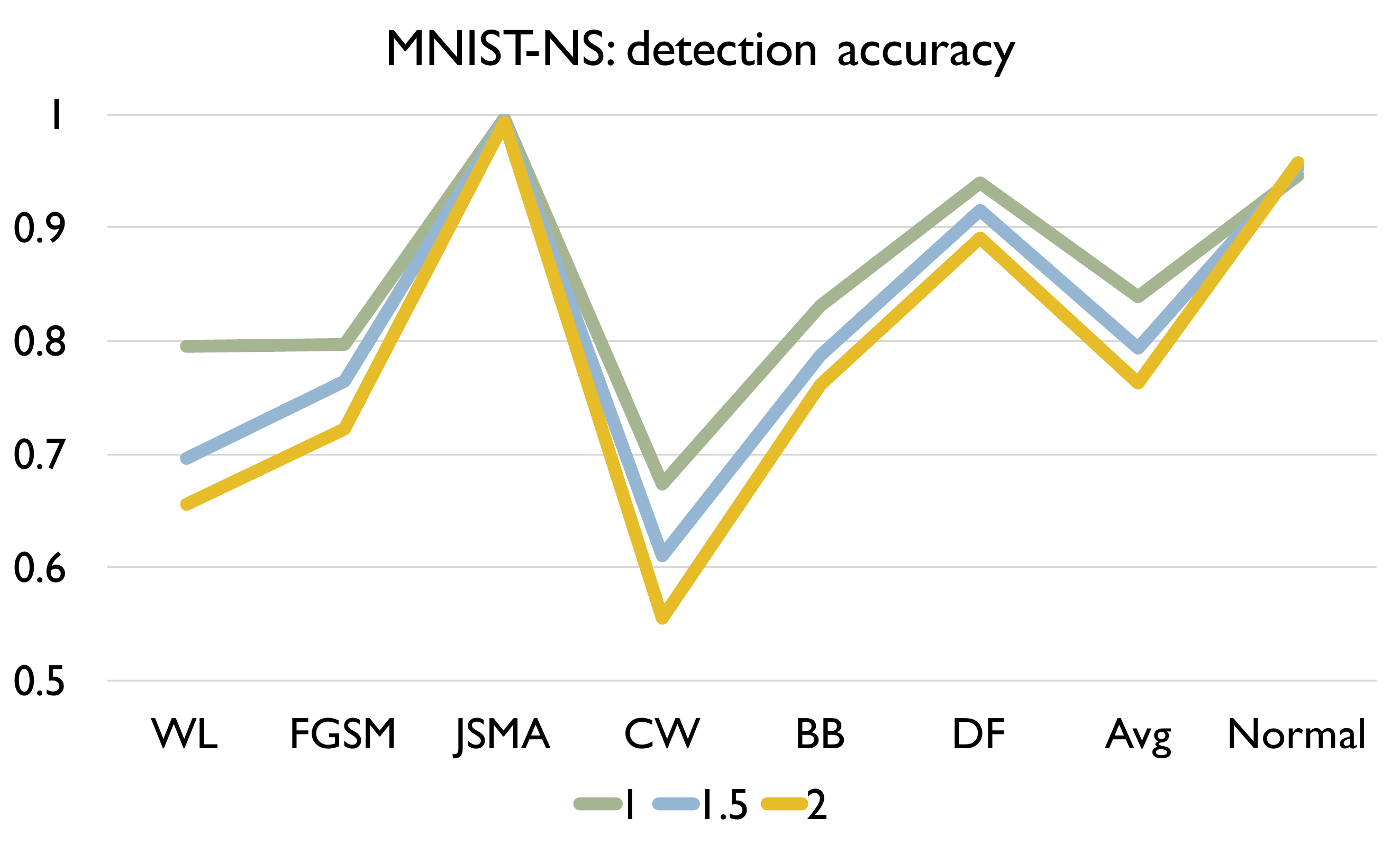}%
\end{subfigure}% 
\begin{subfigure}[b]{.25\textwidth}
   \centering 
   \label{fig:mu:ns}%
   \includegraphics[height=1in]{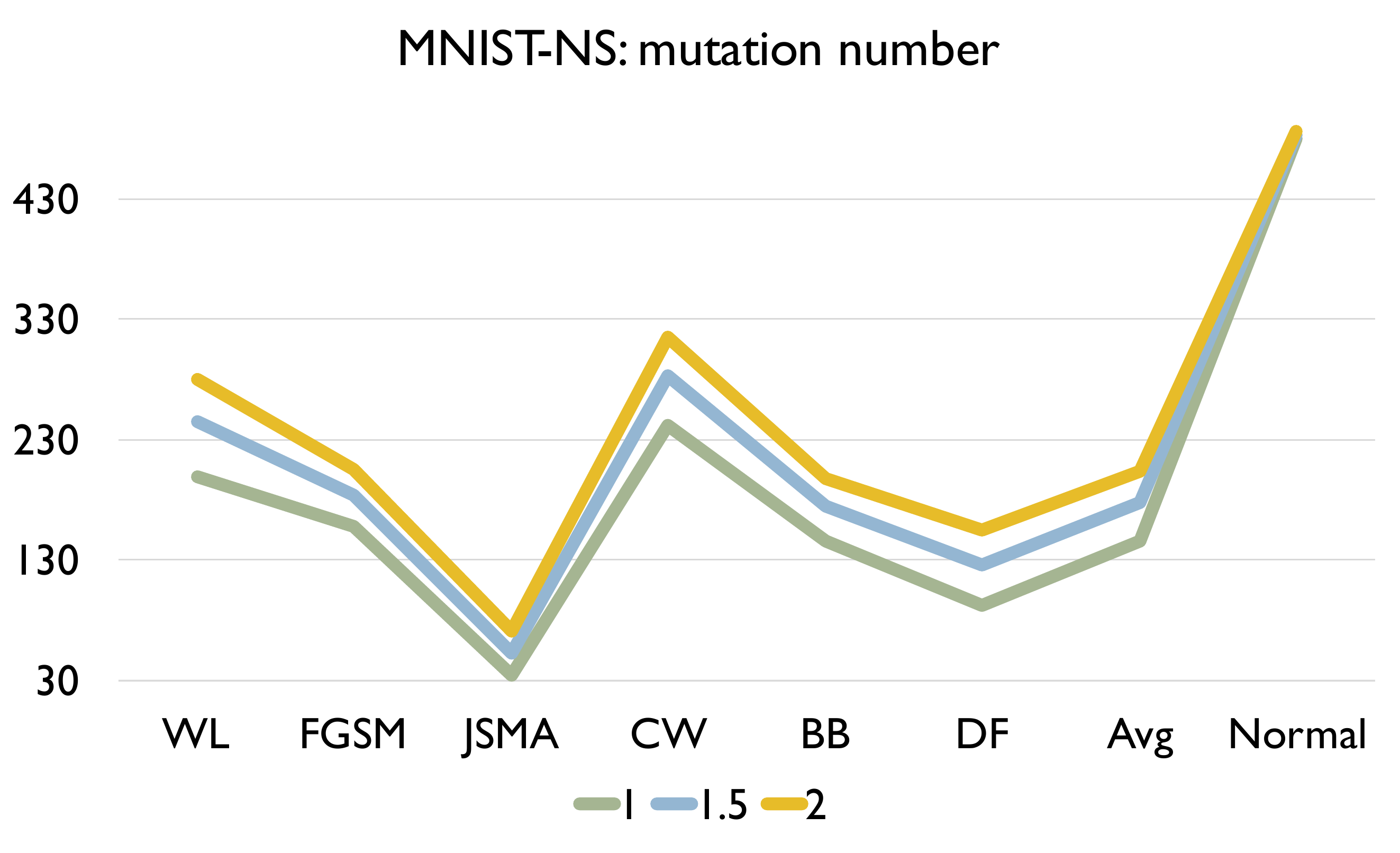}
\end{subfigure}%
\begin{subfigure}[b]{.25\textwidth}
   \centering 
   \label{fig:acc:ws}%
   \includegraphics[height=1in]{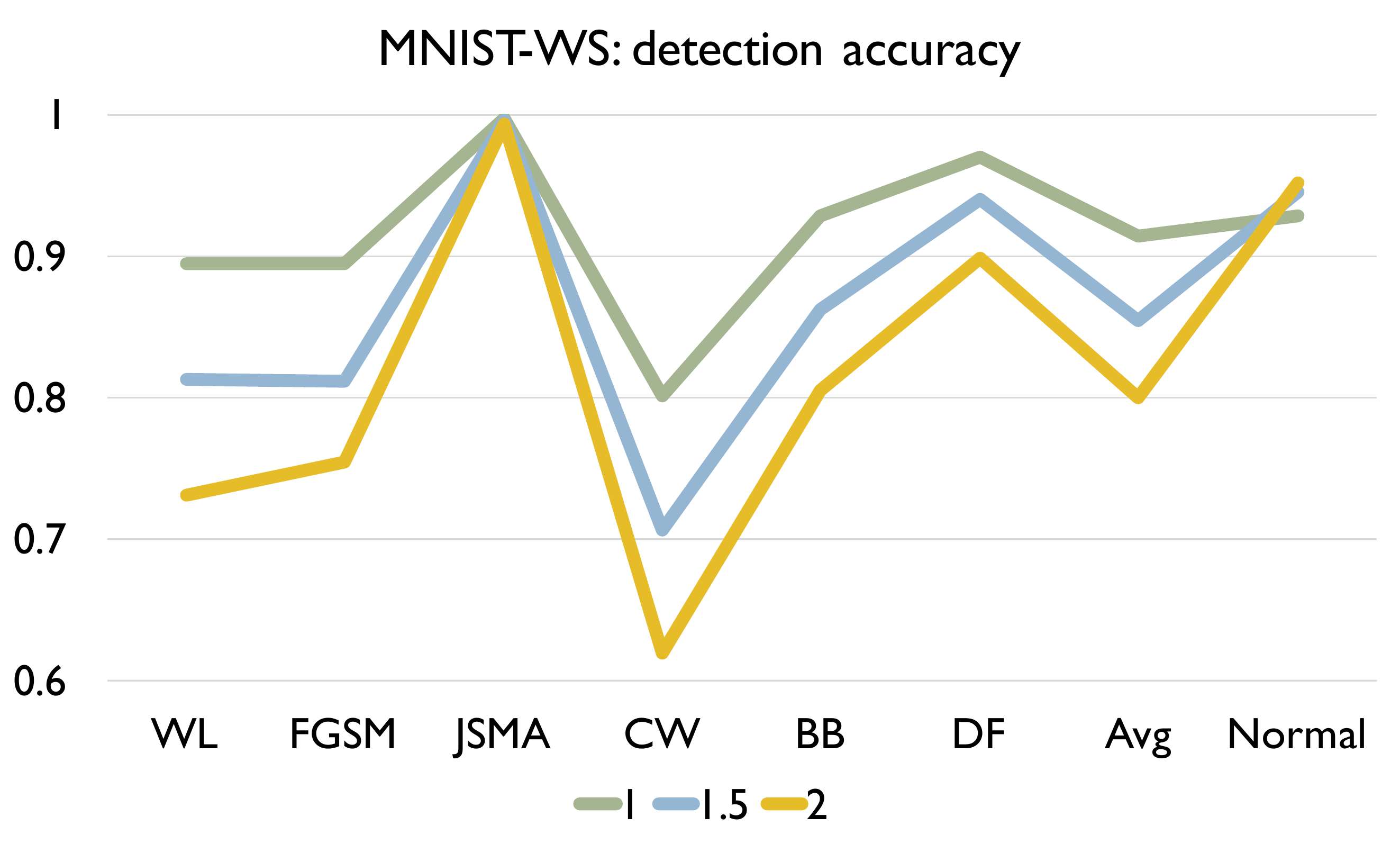}
\end{subfigure}% 
\begin{subfigure}[b]{.25\textwidth}
   \centering 
   \label{fig:mu:ws}%
   \includegraphics[height=1in]{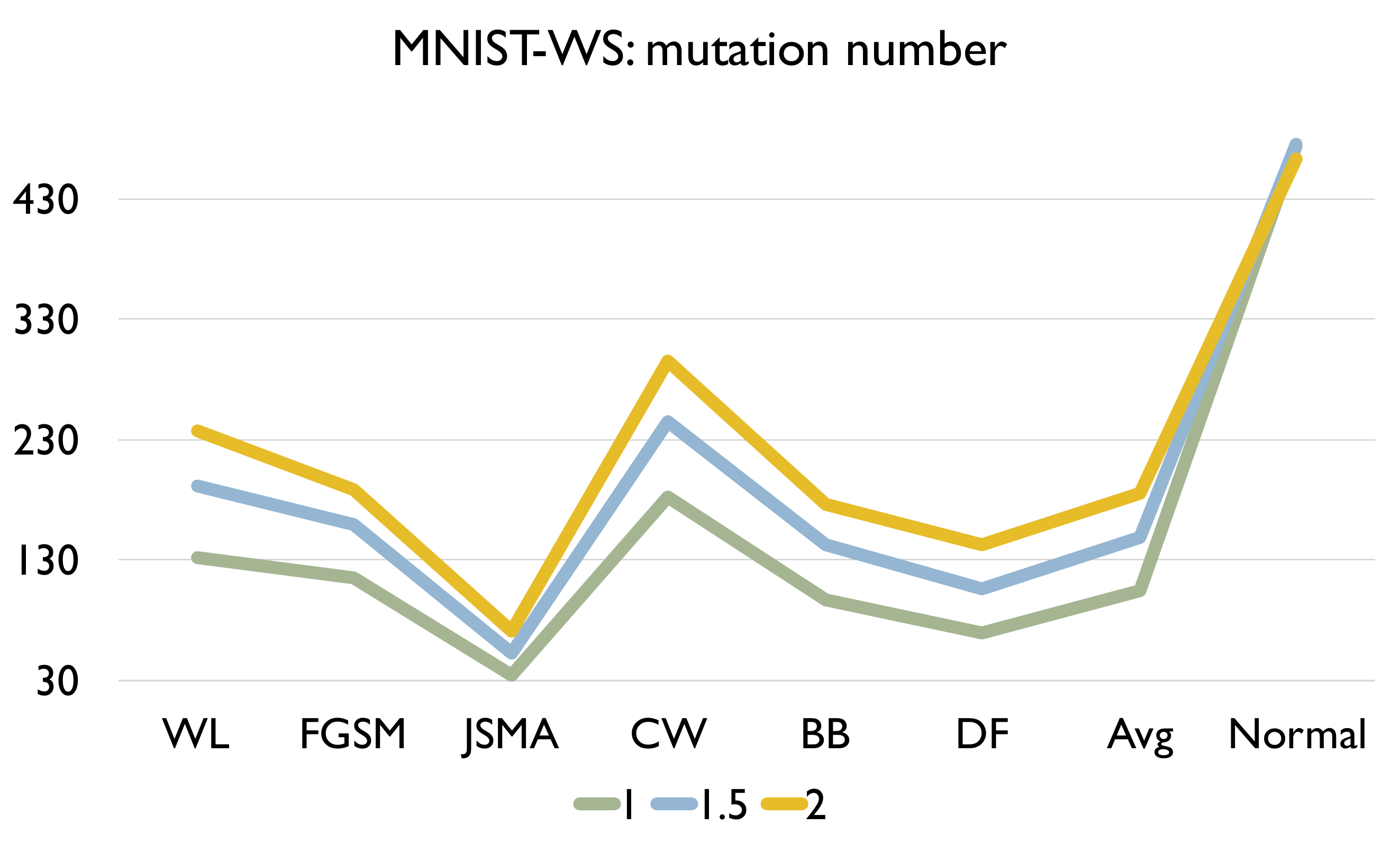}
\end{subfigure}%
\\

\begin{subfigure}[b]{.25\textwidth}
   \centering 
   \label{fig:acc:gf}%
   \includegraphics[height=1in]{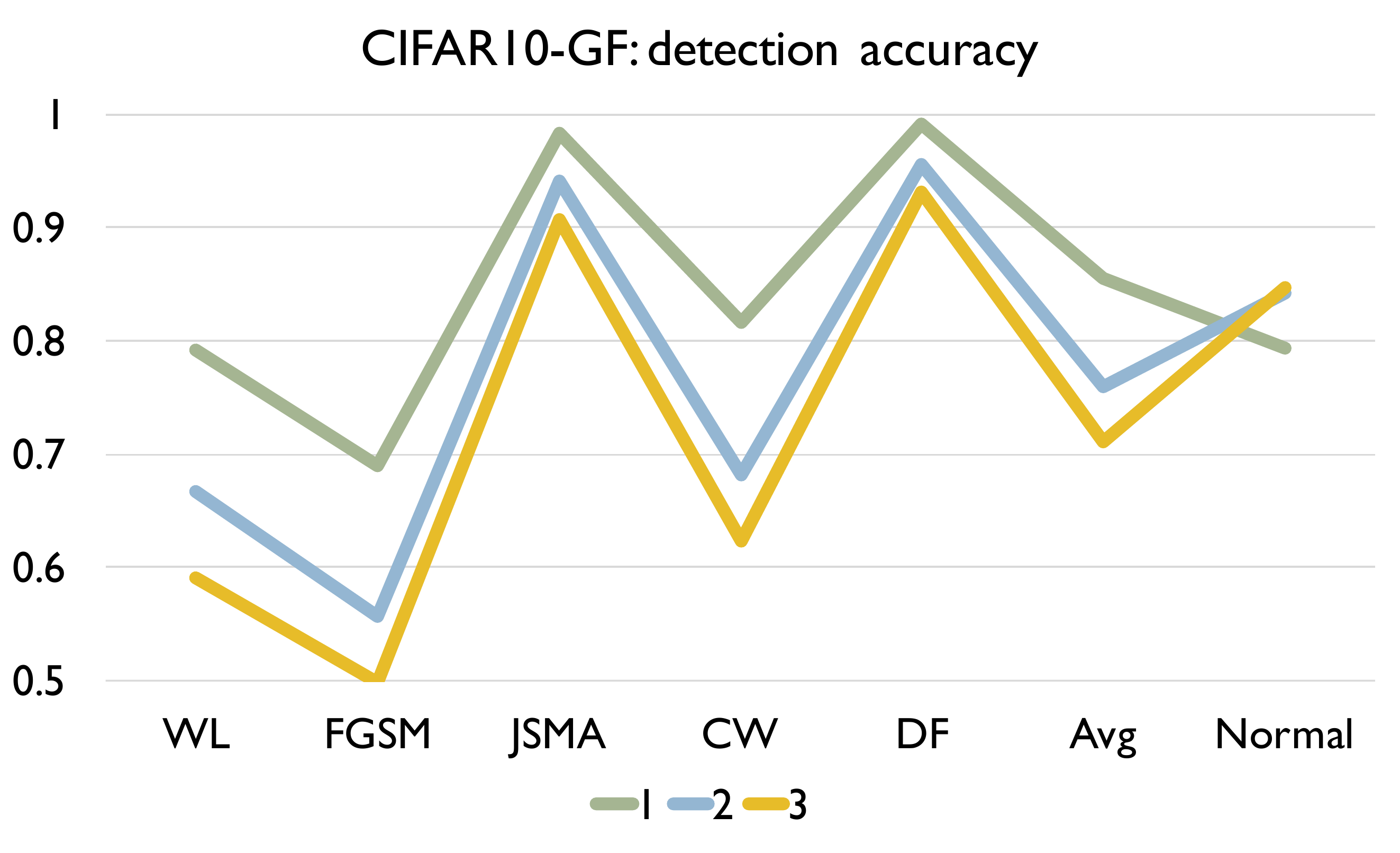}
\end{subfigure}% 
\begin{subfigure}[b]{.25\textwidth}
   \centering 
   \label{fig:mu:gf}%
   \includegraphics[height=1in]{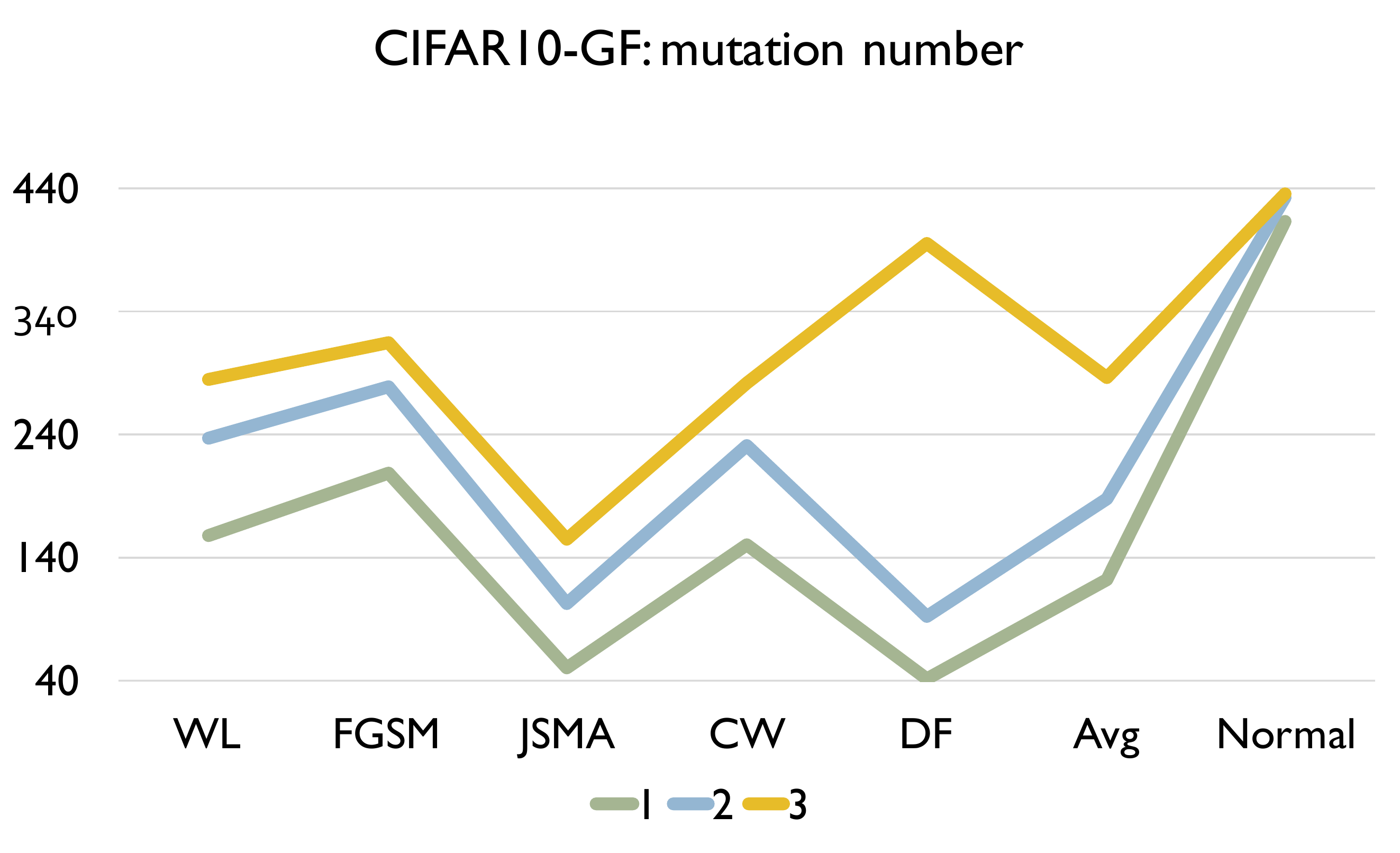}
\end{subfigure}%
\begin{subfigure}[b]{.25\textwidth}
   \centering 
   \label{fig:acc:nai}%
   \includegraphics[height=1in]{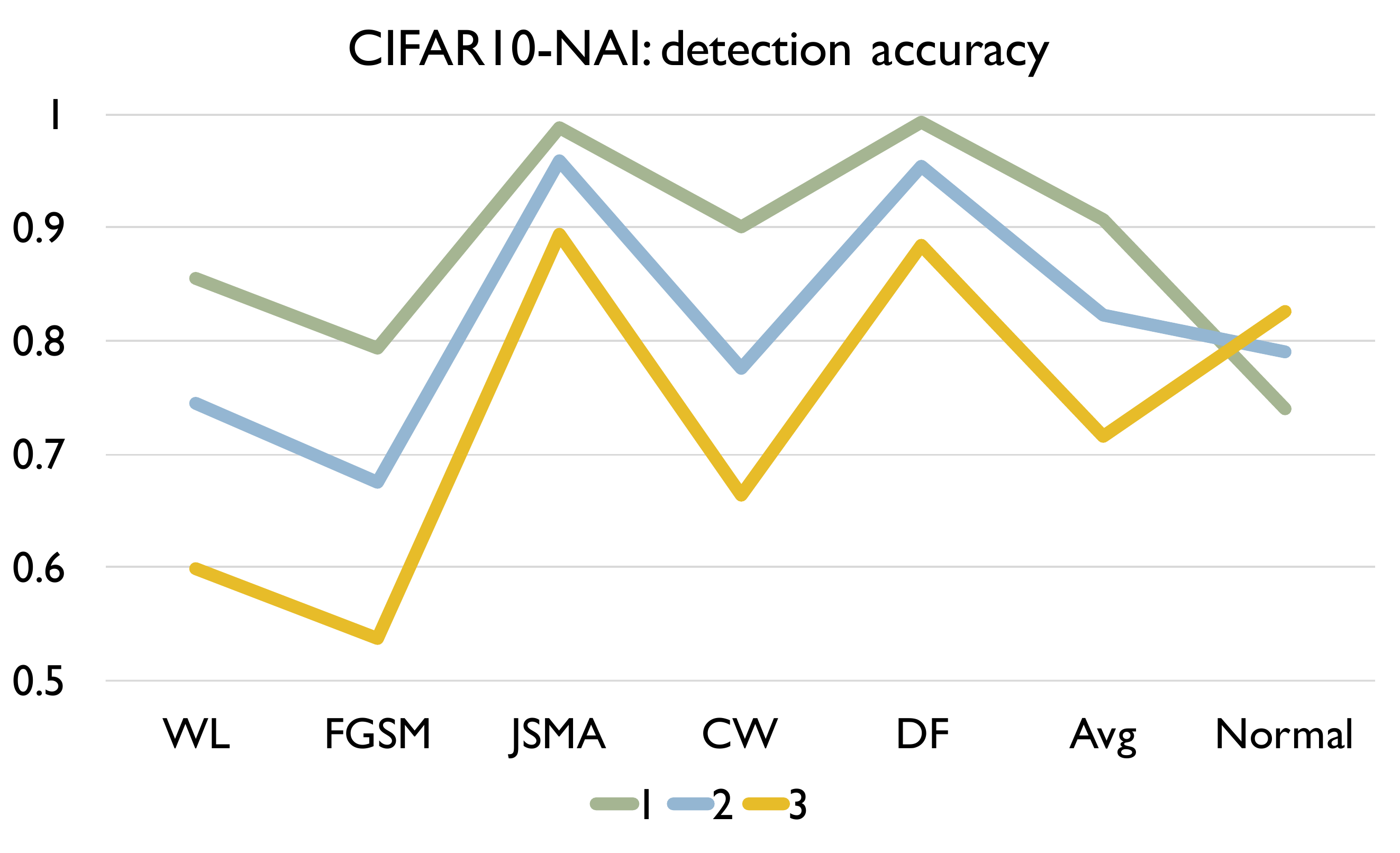}
\end{subfigure}% 
\begin{subfigure}[b]{.25\textwidth}
   \centering 
   \label{fig:mu:nai}%
   \includegraphics[height=1in]{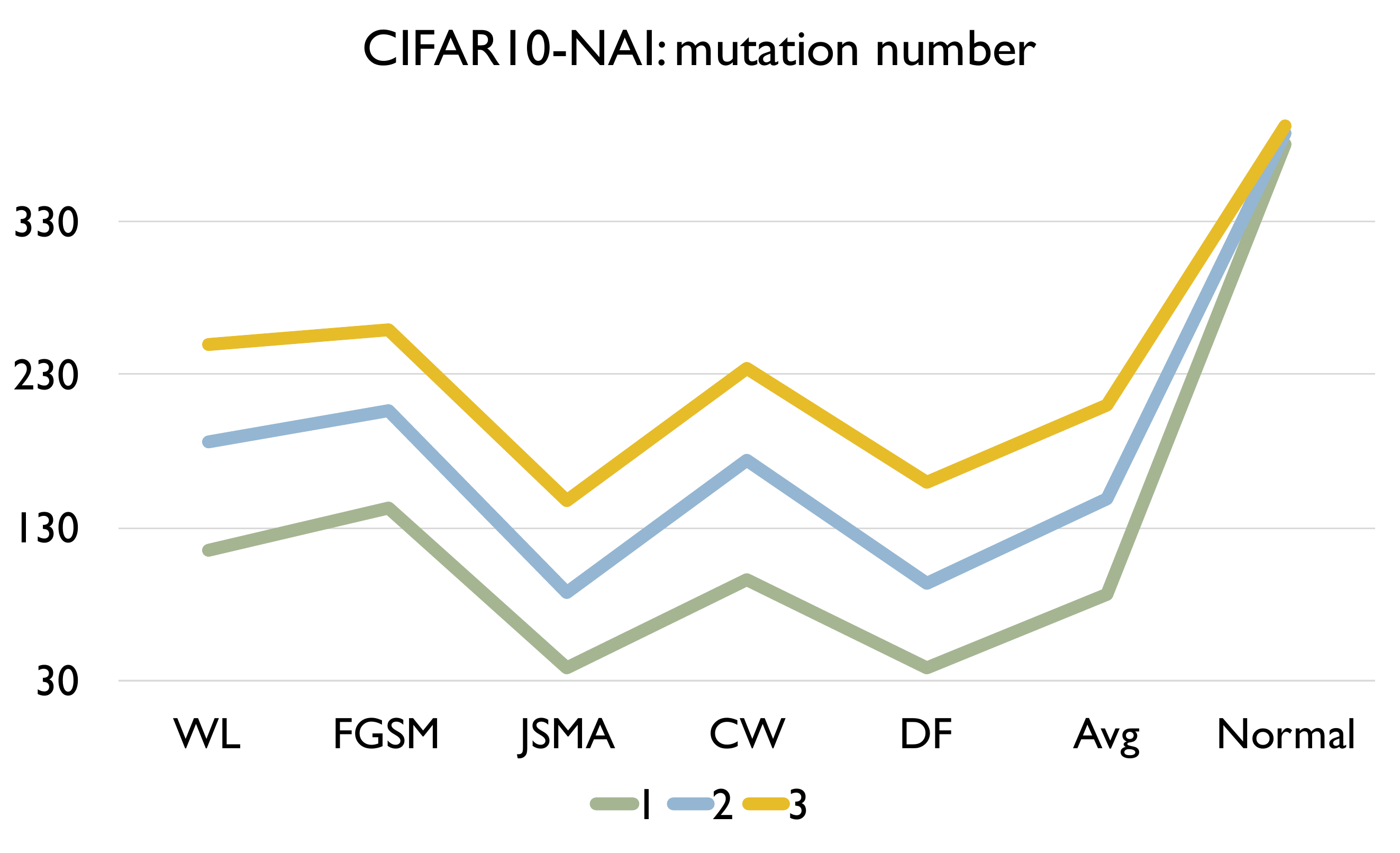}
\end{subfigure}%
\\
\begin{subfigure}[b]{.25\textwidth}
   \centering 
   \label{fig:acc:ns}%
   \includegraphics[height=1in]{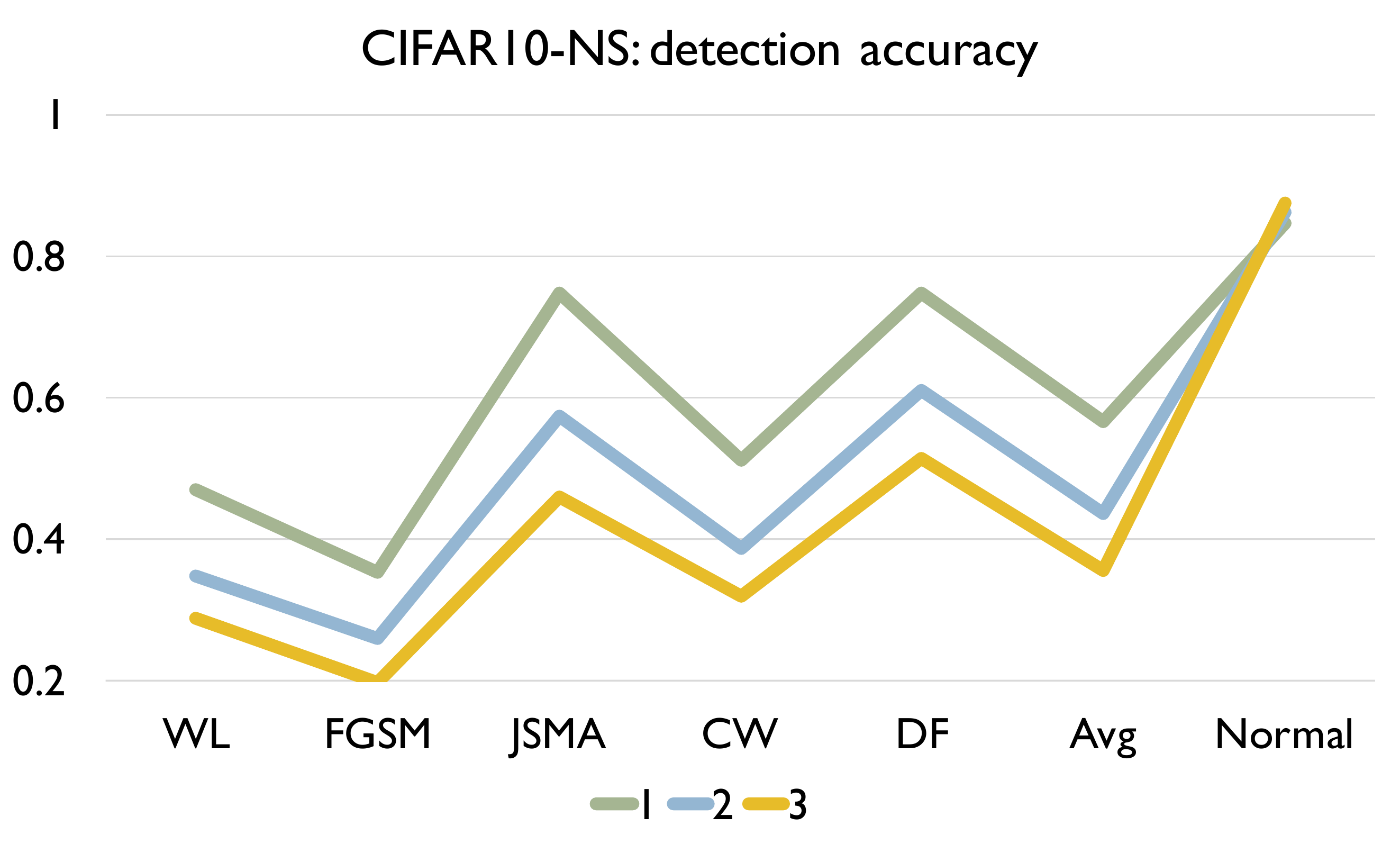}
\end{subfigure}% 
\begin{subfigure}[b]{.25\textwidth}
   \centering 
   \label{fig:mu:ns}%
   \includegraphics[height=1in]{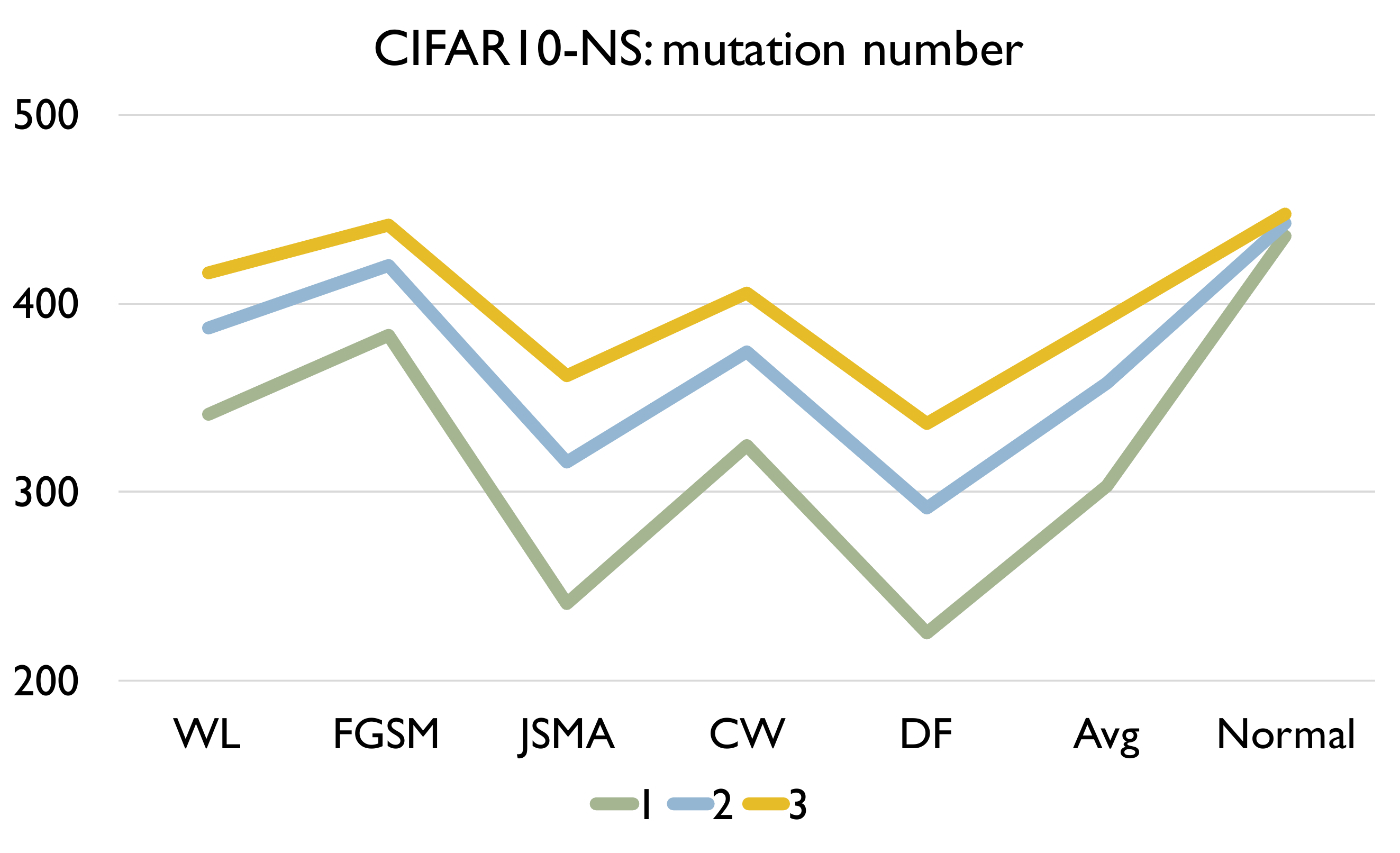}
\end{subfigure}%
\begin{subfigure}[b]{.25\textwidth}
   \centering 
   \label{fig:acc:ws}%
   \includegraphics[height=1in]{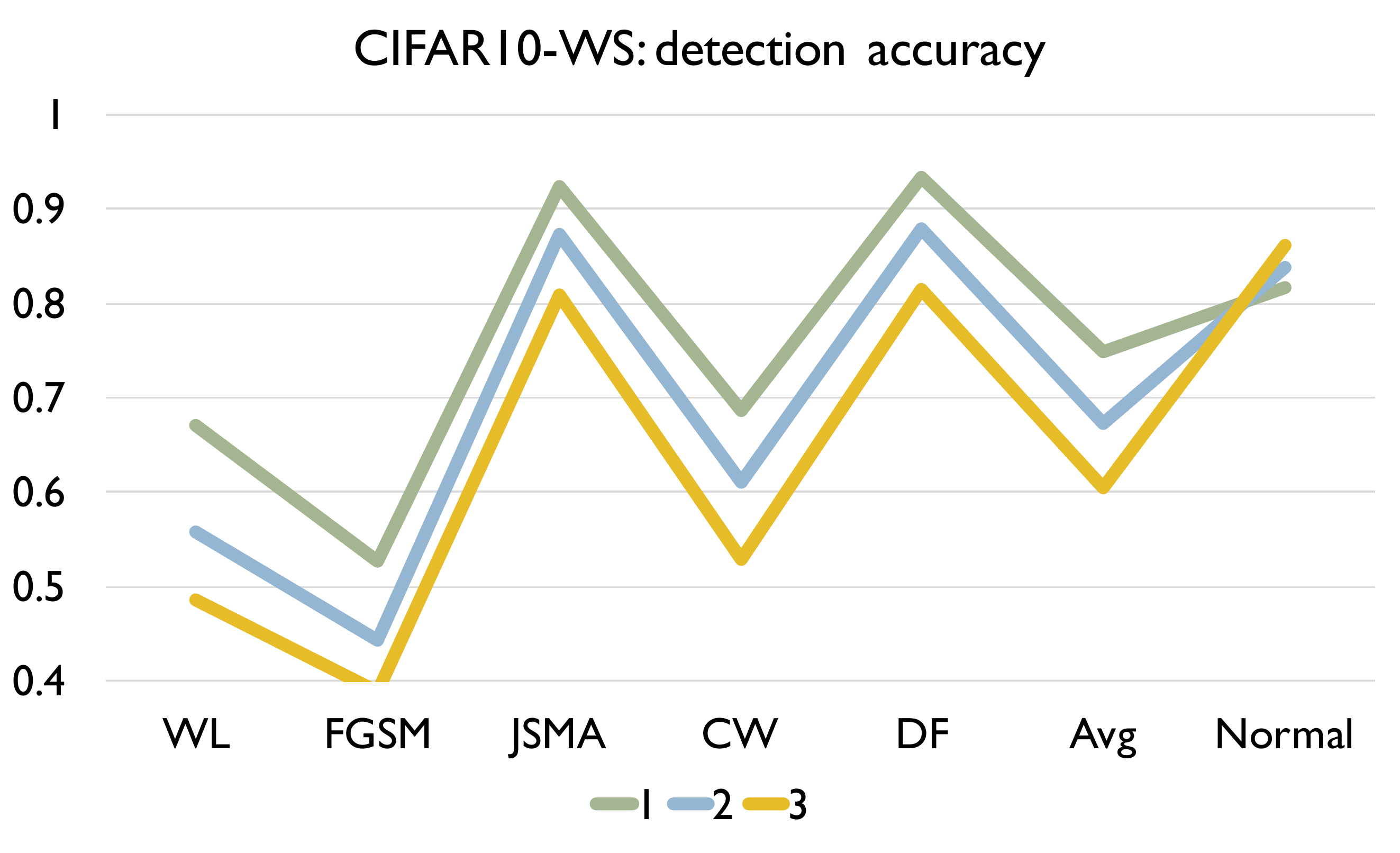}
\end{subfigure}% 
\begin{subfigure}[b]{.25\textwidth}
   \centering 
   \label{fig:mu:ws}%
   \includegraphics[height=1in]{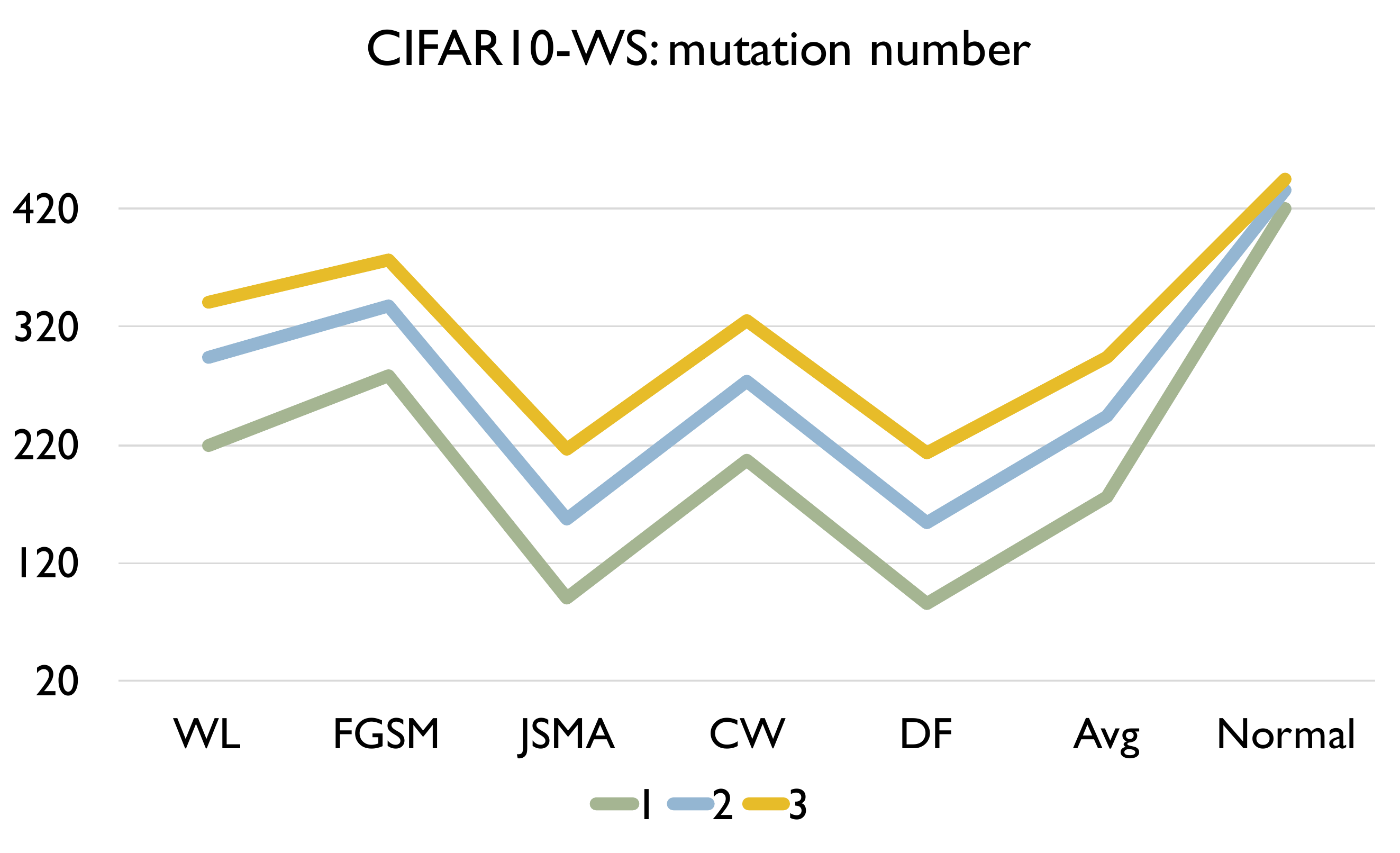}
\end{subfigure}%
   \caption{Detection accuracy and number of mutated models needed.}
\label{fig:acc}
\end{figure*}

\noindent \emph{RQ3: How effective is our detection algorithm based on LCR under model mutation?} To answer the question, we apply our detection algorithm (Algorithm~\ref{alg:sprt}) on each set of adversarial samples generated using each attack and evaluate the accuracy of the detection in Figure~\ref{fig:acc}. We also report the accuracy of our algorithm on a set of normal samples. The results are based on the set of models generated using mutation rate 0.05 for MNIST and 0.005 for CIFAR10 as they have good balance between detecting adversarial and normal samples. 

We set the parameters of Algorithm~\ref{alg:sprt} as follows. Since different kind of attacks have different LCR but the LCR of normal sample is relatively stable, we choose to test against the LCR of normal samples. Specifically, we set the threshold $\varsigma_h$ to be $\rho\cdot\varsigma_{nr}$, where $\varsigma_{nr}$ is the upper bound of the confidence interval of $\varsigma_{nor}$ and $\rho$ ($\ge 1$) is a hyper parameter to control the sensitivity of detecting adversarial samples in our algorithm. The smaller $\rho$ is, the more sensitive our algorithm is to detect adversarial samples. The error bounds for SPRT is set as $\alpha=0.05,\beta=0.05$. The indifference region is set as $0.1\cdot\varsigma_{nr}$. 

Figure~\ref{fig:acc} shows the detection accuracy and average number of model mutants needed for the detection using the 4 mutation operators for MNIST and CIFAR10 dataset respectively. 
% For each operator and dataset, we vary $\rho$ to see its effect. 
We could observe that our detection algorithm achieves high accuracy on every kind of attack for every mutation operator. On average, the GF/NAI/NS/WS operators achieves accuracy of 94.9\%/96.4\%/83.9\%/91.4\% with 75.5/74.1/145.3/105.4 mutated models for MNIST (with $\rho$=1) and 85.5\%/90.6\%/56.6\%/74.8\% (with $\rho$=1) with 121.7/86.1/303/176.2 mutated models for CIFAR10 on detecting the 6 kinds of adversarial samples. Meanwhile, we maintain high detection accuracy of normal samples as well, i.e., 90.8\%/89.7\%/94.7\%/92.9\% for MNIST (with $\rho$=1) and 79.3\%/74\%/84.6\%/81.6\% (with $\rho$=1) for CIFAR10 for the above 4 operators respectively. Notice that for CIFAR10, we could not train a good substitute model (the accuracy is below 50\%) using Black-box attack and thus have no result. The results show that our detection algorithm is able to detect most of adversarial samples effectively. In addition, we observe that the more accurate is the original (and as a result the mutated) DNN model is (e.g., MNIST), the better is our algorithm. Besides, we are able to achieve accuracy close to 1 for JSMA and DF. We also recommend to use NAI/GF operators over NS/WS operators as they have consistently better performance than the others. We thus have the following answer to RQ3.
\begin{framed}
\noindent \emph{Answer to RQ3: Our detection algorithm based on statistical hypothesis testing could effectively detect adversarial samples.}
\end{framed}

\emph{Effect of $\rho$} In this experiment, we vary the hyper parameter $\rho$ to see its effect on the detection. As shown in Figure~\ref{fig:acc}, we set $\rho$ as $\{1,1.5,2\}$ for MNIST and $\{1,2,3\}$ for CIFAR10. We could observe that as we increase $\rho$, we have a lower accuracy on detecting adversarial samples but a higher accuracy on detecting normal samples. The reason is that as we increase $\rho$, the threshold for the detection increases. In this case, our algorithm will be less sensitive to detect adversarial samples since the threshold is higher. We could also observe that we would need more mutations with a higher threshold. In summary, the selection of $\rho$ could be application specific and our practical guide is to set a small $\rho$ if the application has a high safety requirement and vice versa. \\

\noindent \emph{RQ4: What is the cost of our detection algorithm?} The cost of our algorithm mainly consists of two parts, i.e., generating mutated models (denoted by $c_g$) and performing forward propagation (denoted by $c_f$) to obtain the label of an input sample by a DNN model. The total cost of detecting an input sample is thus $C=n\cdot(c_g+c_f)$, where $n$ is the number of mutants needed to draw a conclusion based on Algorithm~\ref{alg:sprt}. 

We estimate $c_f$ by performing forward propagation for 10000 images on a MNIST and CIFAR10 model respectively. The detailed results are shown in Tabel~\ref{tb:cost}. Note that $c_g$ is the time used to generate an accurate model (retaining at least 90\% accuracy of the original model) and the cost to generate an arbitrary mutated model is much less.
In practice, we could generate and cache a set of mutated models for the detection of a set of samples. Given a set of $m$ samples, the total cost for the detection is reduced to $C(m)=m\cdot n\cdot c_f+n*c_g$. 
In practice, our algorithm could detect an input sample within 0.1 second (with cached models) using a single machine. We remark that our algorithm can be parallelized easily by evaluating a set of models at the same time which would reduce the cost significantly. We thus have the following answer to RQ4.
\begin{framed}
\noindent \emph{Answer to RQ4: Our detection algorithm is lightweight and easy to parallel.}
\end{framed}

% cifar10:5e-3p, 1.02,0.2
% mnist 5e-2p, 1.2,0.2
\begin{table}[t]
\caption{Cost analysis of our algorithm.}
\centering
\begin{tabular}{@{}ccccc@{}}
\toprule
Dataset & $c_f$ & operator & $c_g$ & $n$\\ \midrule
\multirow{4}{*}{MNIST}   &  0.7 ms    & NAI    &6.191 s  & 68.7789    \\
        &     0.5 ms & NS       & 6.336 s    & 173.0040 \\
        &    0.3 ms   & WS       & 7.657 s    & 107.6702 \\
        &    0.3 ms & GF       & 1.398 s  & 91.1747\\ \midrule
\multirow{4}{*}{CIFAR10}   & 0.3 ms      & NAI      & 16.101 s    &69.0873\\
        &  0.5 ms      & NS       & 9.475 s     &283.9628\\
        &  0.4 ms & WS       & 9.251 s    &165.6373\\
        &  0.4 ms    & GF       & 11.894 s    & 127.2767 \\ \bottomrule
\end{tabular}
\label{tb:cost}
\end{table}

\subsection{Threats to Validity}

% Despite the effectiveness and efficiency of our approach, there are some threats to validity summarized as follows. 

First, our experiment is based on a limited set of test subjects so far. Our experience is that the more accurate the original model and the mutated models are, the more effective and more efficient our detection algorithm is. The reason is that the LCR distance between adversarial samples and normal samples will be larger if the model is more accurate, which is good for our detection. In some applications, however, the accuracy of the original models may not be high.  
Secondly, the detection algorithm will have some false positives. Since our detection algorithm is threshold-based, there will be some false alarms along with the detection. Meanwhile, there is a tradeoff between avoiding false positives or false negatives as discussed above (i.e., in the selection of $\rho$). Thirdly, the detection of normal samples typically needs more mutations. The reason is that we choose to test against $\varsigma_{nor}$ since we do not know $\varsigma_{adv}$ for an unknown attack. Since normal samples have lower LCR under mutated models in general, they would need more mutations than adversarial samples to draw a conclusion.

\section{Related works}\label{sec:re}
This work is related to studies on adversarial sample generation, detection and prevention. There are several lines of related work in addition to those discussed above.
% Note that the problem of robustness of Deep Neural Networks (DNN) against adversarial examples is a research problem which is emerging rapidly and attracts attentions from multiple communities including machine learning, software engineering and security. 
% Due to the fast growing interests in the area, we will not cover all of them but only list several lines of work as follows.\\

\paragraph{Adversarial training} The key idea of adversarial training is to augment training data with adversarial samples to improve the robustness of the trained DNN itself. Many attack strategies have been invented recently to effectively generate adversarial samples like DeepFool~\cite{DeepFool}, FGSM~\cite{FGSM}, C\&W~\cite{CW}, JSMA~\cite{JSMA}, black-box attacks~\cite{bb} and others~\cite{su2018attacking,sharif2016accessorize,galloway2017attacking,brendel2017decision,xiao2018spatially}. However, adversarial training in general may overfit to the specific kinds of attacks which generate the adversarial samples for training~\cite{madry2017towards} and thus can not guarantee robustness on new kinds of attacks. 

\paragraph{Adversarial sample detection} Another direction is to automatically detect those adversarial samples that a DNN will mis-classify. One way is to train a `detector' subnetwork from normal samples and adversarial samples~\cite{detection_subnn}. Alternative detection algorithms are often based on the difference between how an adversarial sample and a normal sample would behave in the softmax output~\cite{feature_squezze,detection1,detection2,feinman2017detecting} or under random perturbations~\cite{detection_ours}. 

\paragraph{Model robustness} Different metrics has been proposed in the machine learning community to measure and provide evidence on the robustness of a target DNN~\cite{xu2012robustness,weng2018evaluating}. Besides, in~\cite{deepxplore} and the following work~\cite{dnn_test,ma2018deepgauge}, neuron coverage and its extensions are argued to be the key indicators of the DNN robustness. In~\cite{measure_cons}, they proposed adversarial frequency and adversarial severity as the robustness metrics and encode robustness as a linear program.  

\paragraph{Testing and formal verification} Testing strategies including white-box~\cite{deepxplore,deeptest}, black-box~\cite{feature_guided_test} and mutation testing~\cite{deepmutation} have been proposed to generate adversarial samples more efficiently for adversarial training. However, testing can not provide any safety guarantee in general. There are also attempts to formally verify certain safety properties against the DNN to provide certain safety guarantees~\cite{safety_verify,reluplex,prove_robust,weng2018towards}. 
% These methods are however too computationally expensive and not scalable to large DNN.

\section{Conclusion}\label{sec:con}

In this work, we propose an approach to detect adversarial samples for Deep Neural Networks at runtime. Our approach is based on the evaluated hypothesis that most adversarial samples are much more sensitive to model mutations than normal samples in terms of label change rate. We then propose to detect whether an input sample is likely to be normal or adversarial by statistically checking the label change rate of an input sample under model mutations. We evaluated our approach on MNIST and CIFAR10 datasets and showed that our algorithm is both accurate and efficient to detect adversarial samples.  

\section*{Acknowledgment}
Xinyu Wang is the corresponding author. This research was supported by grant RTHW1801 in collaboration with the Shield Lab of Huawei 2012 Research Institute, Singapore. We are thankful to the discussions and feedbacks from them. This research was also partially supported by the National Basic Research Program of China (the 973 Program) under grant 2015CB352201 and NSFC Program (No. 61572426).

% \fi
\IEEEpeerreviewmaketitle


% \section*{Acknowledgment}




% http://www.michaelshell.org/tex/bibtex/
%\bibliographystyle{IEEEtran}


% \end{thebibliography}

% \clearpage
\bibliographystyle{plain}
\bibliography{main}

\end{document}